\documentclass{article}

\PassOptionsToPackage{numbers, compress}{natbib}
\usepackage[preprint]{neurips_2026}

\usepackage[utf8]{inputenc} 
\usepackage[T1]{fontenc}    
\usepackage{amsmath}        
\usepackage{graphicx}       
\usepackage{array}          
\usepackage{adjustbox}      
\usepackage{float}          
\usepackage{makecell}       
\usepackage{multirow}       
\usepackage[inline]{enumitem}
\usepackage{url}            
\usepackage{booktabs}       
\usepackage{amssymb}        
\usepackage{nicefrac}       
\usepackage{microtype}      
\usepackage[table]{xcolor}         
\usepackage{booktabs}
\usepackage{tabularx}
\IfFileExists{tcolorbox.sty}{%
  \usepackage{tcolorbox}
  \tcbset{
    visprompt/.style={
      colback=gray!3,
      colframe=black!45,
      boxrule=0.4pt,
      arc=1pt,
      left=4pt,
      right=4pt,
      top=4pt,
      bottom=4pt,
      fonttitle=\bfseries
    }
  }
}{%
  \newenvironment{tcolorbox}[1][]{%
    \par\smallskip\noindent
    \begin{minipage}{0.96\linewidth}
    \hrule\smallskip
  }{%
    \smallskip\hrule
    \end{minipage}
    \par\smallskip
  }
}
\newif\ifvisTikz
\IfFileExists{tikz.sty}{\usepackage{tikz}\visTikztrue}{\visTikzfalse}
\usepackage{hyperref}       

\newcommand{\tick}{\textcolor{green!60!black}{\checkmark}}
\newcommand{\cross}{\textcolor{red!70!black}{\ensuremath{\times}}}
\newcommand{\ie}{i.e.}

\title{VisPhyWorld: Probing Physical Reasoning via Code-Driven Video Reconstruction}

\author{%
  Jiarong Liang\textsuperscript{1,*} \quad
  Max Ku\textsuperscript{1,*} \quad
  Ka-Hei Hui\textsuperscript{2} \quad
  Ping Nie\textsuperscript{3} \quad
  Wenhu Chen\textsuperscript{1} \\
  \normalfont\small
  \textsuperscript{1}University of Waterloo \quad
  \textsuperscript{2}Autodesk AI Lab \quad
  \textsuperscript{3}Independent Researcher
}

\makeatletter
\renewcommand{\@noticestring}{%
  Preprint. \textsuperscript{*}Equal contribution. Correspondence to
  \texttt{jiarongliangcs@gmail.com}, \texttt{m3ku@uwaterloo.ca} and \texttt{wenhu.chen@uwaterloo.ca}.%
}
\makeatother

\begin{document}

\maketitle

\begin{abstract}
Evaluating whether Multimodal Large Language Models (MLLMs) genuinely reason about physical dynamics remains challenging.
Most existing benchmarks rely on recognition-style protocols such as Visual Question Answering (VQA) and Violation of Expectation (VoE), which can often be answered without committing to an explicit, testable physical hypothesis.
We propose \textbf{VisPhyWorld}, an execution-based framework that evaluates physical reasoning by requiring models to generate executable simulator code from visual observations. 
By producing runnable code, the inferred world representation is directly inspectable, editable, and falsifiable. This separates physical reasoning from rendering.
Building on this framework, we introduce \textbf{VisPhyBench}, comprising 209 evaluation scenes derived from 108 physical templates and a systematic protocol that evaluates how well models reconstruct appearance and reproduce physically plausible motion. Our pipeline produces valid reconstructed videos in 97.7\% of benchmark runs before fallback. Experiments show that while state-of-the-art MLLMs achieve strong semantic scene understanding, they struggle to accurately infer physical parameters and to simulate consistent physical dynamics. Our code is available https://github.com/TIGER-AI-Lab/VisPhyWorld
\end{abstract}

\begin{figure}[htbp]
  \centering
  \begin{minipage}[c]{0.62\columnwidth}
    \centering
    \includegraphics[width=\linewidth]{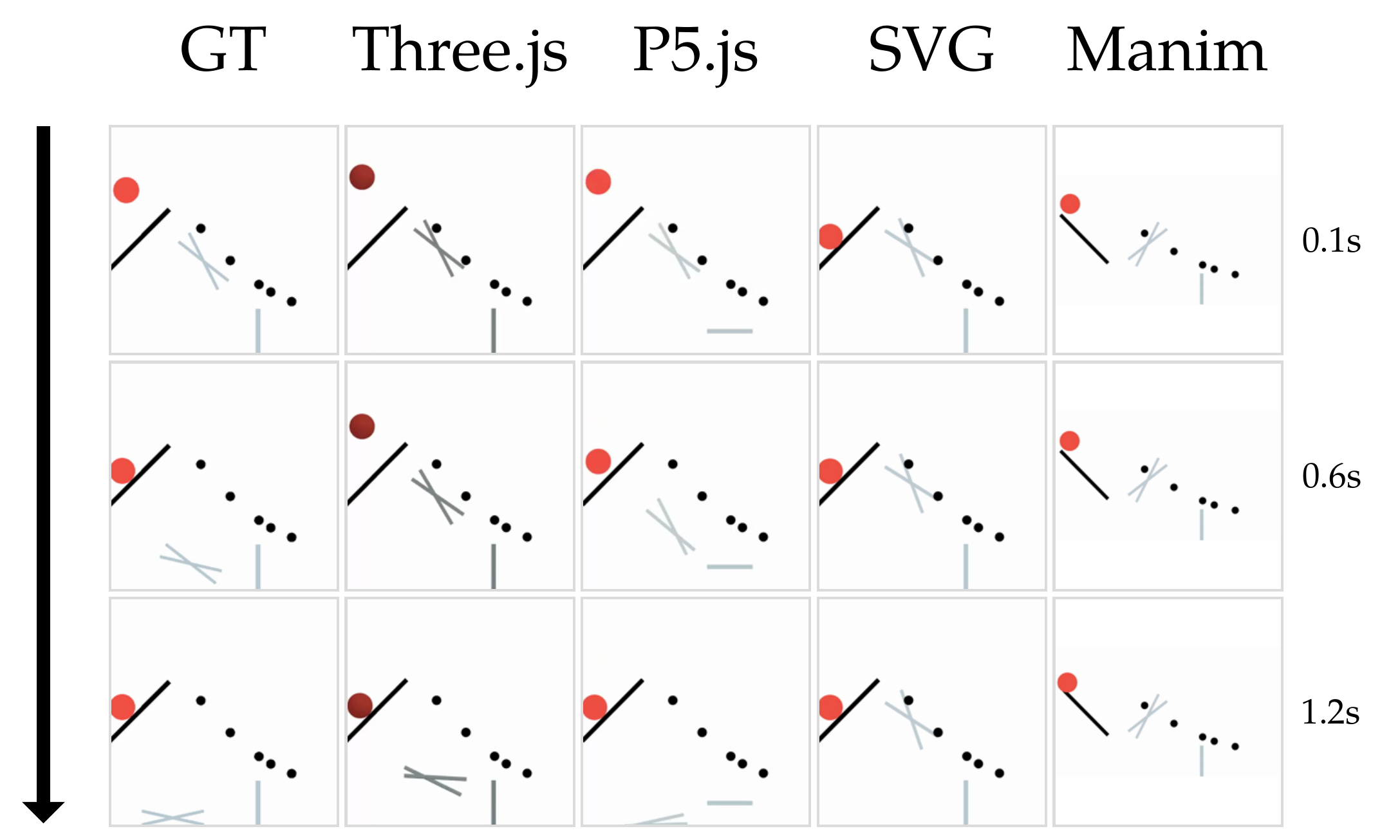}
  \end{minipage}
  \hfill
  \begin{minipage}[c]{0.34\columnwidth}
    \caption{\textbf{MLLMs struggle to simulate physical dynamics.} Under the same inputs, rigid-body backends (Three.js/P5.js) produce more physically consistent rollouts, whereas non-physics backends (SVG/Manim) often exhibit implausible motion or contact artifacts such as interpenetration.}
    \label{fig:engine_eval_task10010}
  \end{minipage}
\end{figure}

\section{Introduction}
\label{sec:introduction}

\begin{figure*}[ht]
    \centering
    \includegraphics[width=0.9\textwidth]{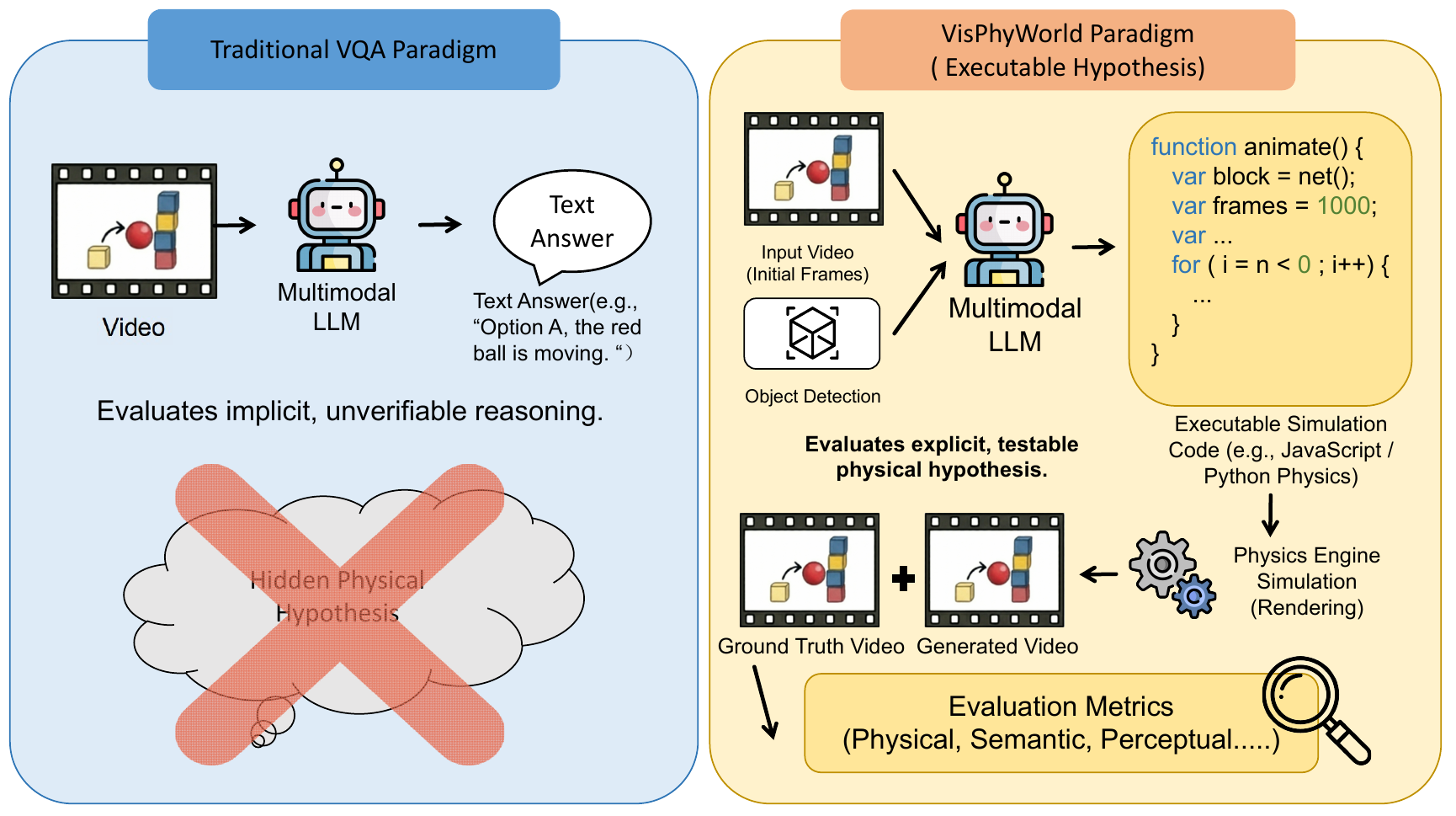}
    
    \caption{Unlike traditional VQA paradigms, \textbf{VisPhyWorld accesses physical understanding evaluation} by requiring MLLMs to actively reconstruct scenes via executable code, offering superior reasoning explainability compared to traditional paradigms.}
    \label{fig:teaser}
\end{figure*}

\begin{figure*}[ht]
    \centering
    \includegraphics[width=0.9\textwidth]{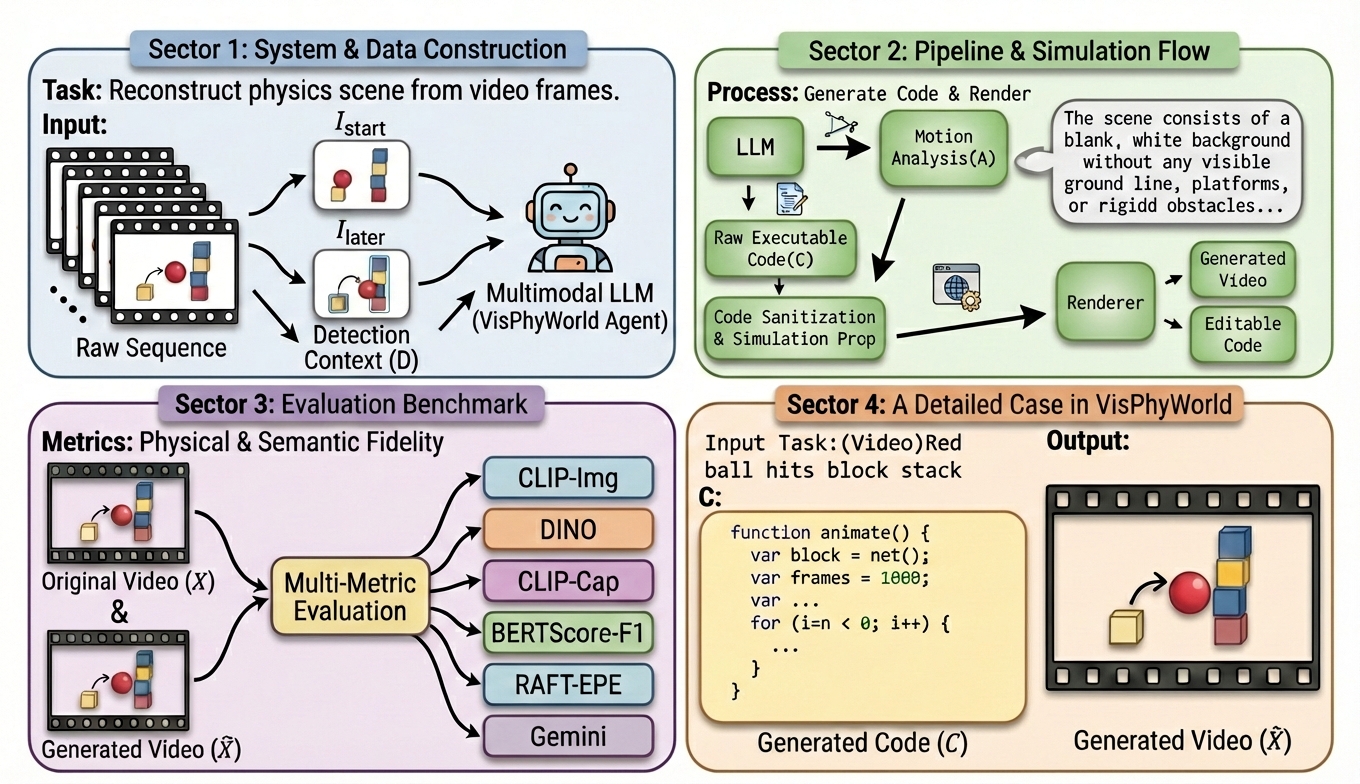}
    
    \caption{\textbf{VisPhyWorld Framework.}  
    \textbf{(1) System \& Data Construction:} We process raw video sequences to extract key frames ($I_{\text{start}}, I_{\text{later}}$) and detection contexts using multimodal agents. 
    \textbf{(2) Pipeline \& Simulation Flow:} An LLM-based agent performs motion analysis and generates raw executable code, which is then sanitized and rendered. 
    \textbf{(3) Evaluation Benchmark:} We propose a multi-metric benchmark integrating semantic and physical fidelity to compare generated videos $\hat{X}$ with ground truth $X$. 
    \textbf{(4) A Detailed Case:} An example illustrating how VisPhyWorld translates a collision scene (red ball hits block stack) into executable simulation logic.}
    \label{fig:framework}
\end{figure*}
Recent advances in Multimodal Large Language Models (MLLMs) have led to impressive performance on a wide range of visual and language tasks~\citep{fu2024mmesurveycomprehensivesurveyevaluation}. However, assessing whether such models exhibit principled physical reasoning remains challenging. Existing evaluation protocols often rely on recognition-based queries or surface-level judgments, which can obscure whether correct outputs arise from coherent physical reasoning or from learned visual correlations~\citep{chen2023theoremqa, shen2025phyxdoesmodelwits}. Most benchmarks probe physical understanding through passive recognition tasks such as Visual Question Answer-
ing (VQA)-style or Violation of Expectation (VoE)-inspired recognition tasks (e.g. CLEVRER~\citep{yi2020clevrercollisioneventsvideo}, GRASP~\citep{jassim2024graspnovelbenchmarkevaluating}, MVPBench~\citep{krojer2025shortcutawarevideoqabenchmarkphysical})). These settings can reward dataset-driven guessing, encouraging memorized priors and surface-level pattern matching rather than genuine causal physical understanding~\citep{
pezeshkpour2024large, keluskar2024llms}. This challenge is particularly acute for MLLMs, which typically output only text and therefore do not provide predictive likelihoods or measures of surprise commonly used to evaluate generative world models~\citep{garrido2025intuitive}. 
We therefore argue that in this context, \textbf{evaluation should require reconstruction and re-simulation, forcing models to commit to an explicit physical hypothesis rather than merely select an answer or text reasoning.} We propose VisPhyWorld, a paradigm shift: using executable code as a test of physical understanding, as illustrated in Figure ~\ref{fig:teaser}. VisPhyWorld probes the physical reasoning capabilities of MLLMs through visual-to-code reconstruction. Given two key frames (and optionally object detections), the model produces executable simulation code that recreates the scene and rolls it forward to synthesize future frames as shown in Figure~\ref{fig:framework}. This process not only produces the video but does so in a fully interpretable and editable manner. Beyond the rendered video, VisPhyWorld exposes the generated code itself as a reasoning artifact, making the model’s physical logic directly inspectable.
%

%
We also introduce VisPhyBench, a standardized evaluation suite with a systematic protocol that assesses how well models reconstruct appearance and reproduce physically plausible motion across complementary perspectives. Our investigation reveals a critical insight: while current state-of-the-art LLMs excel at semantic recognition, they exhibit significant limitations in fine-grained physical comprehension, often failing to parameterize simple Newtonian dynamics correctly even in a simple 2D setting, let alone in 3D environments. In summary, our contributions are threefold:
\begin{enumerate}
    \setlength{\itemsep}{0pt}
    \setlength{\parsep}{0pt}
    \setlength{\topsep}{0pt}
    \renewcommand{\labelenumi}{(\textbf{\arabic{enumi}})}
    \item We propose \textbf{VisPhyWorld}, a framework that uses LLMs to interpret raw video frames and generate executable simulation code for predicting future motion. To our knowledge, this is the first paradigm that evaluates physical reasoning in MLLMs through code reconstruction and re-simulation. By making object states and dynamics explicit, VisPhyWorld provides a direct and interpretable view of a model’s physical understanding.
    
    \item We introduce \textbf{VisPhyBench}, a unified evaluation protocol comprising 209 scenes derived from 108 physical templates that assesses physical understanding through the lens of code-driven resimulation in both 2D and 3D scenes, integrating metrics from different aspects.

    \item We provide an in-depth analysis of current MLLMs, demonstrating that despite their linguistic prowess, they \textbf{fail to grasp the fundamental dynamics of real-world motion}. Our results reveal a critical gap: while models can accurately describe scene contents, they struggled to reconstruct the scene in a way that conformed to the laws of physics, indicating that they rely on superficial visual pattern matching rather than a grounded understanding of physical causality.
\end{enumerate}

\section{Related Work}

\textbf{Intuitive physics.} Understanding the world is commonly studied through physical reasoning tasks that probe models' ability to infer object dynamics, interactions, and causal relationships from visual input~\citep{melnik2023benchmarksphysicalreasoningai, fung2025embodied}. Inspired by findings from developmental psychology showing that infants exhibit sensitivity to physical violations~\citep{BAILLARGEON1985191}, prior work on intuitive physics investigates whether models can anticipate physically plausible outcomes from visual observations. This has been studied through video prediction benchmarks that evaluate the consistency of predicted future dynamics, as well as Violation-of-Expectation (VoE) paradigms~\citep{riochet2018intphys, margoni2024violation, jassim2024graspnovelbenchmarkevaluating}, which assess whether physically implausible events elicit higher predictive surprise. These approaches are well suited to generative world models with explicit prediction objectives. However, they do not naturally extend to MLLMs, which primarily produce textual outputs rather than predictive distributions and therefore cannot be evaluated using likelihood-based or generative video protocols~\citep{garrido2025intuitive}.

Efforts on several datasets and benchmarks have been made~\citep{rajani2020espritexplainingsolutionsphysical, yi2020clevrercollisioneventsvideo, baradel2020cophycounterfactuallearningphysical}, including Phyre~\citep{bakhtin2019phyre, li2024iphyreinteractivephysicalreasoning}, Physion~\citep{bear2021physion, tung2023physionevaluatingphysicalscene}, and IntPhys~\citep{riochet2018intphys, bordes2025intphys}, have been proposed to evaluate intuitive physics using videos generated from physics engines. More recent benchmarks such as PhysicsIQ~\citep{motamed2025generativevideomodelsunderstand_physicsiq}, PhyGenBench~\citep{meng2024worldsimulatorcraftingphysical}, and WorldModelBench~\citep{li2026worldmodelbench} extend this setting to generative video models, focusing on whether predicted videos exhibit physically plausible and temporally consistent dynamics. In parallel, researchers have developed MLLM–based evaluators~\citep{motamed2025travlrecipemakingvideolanguage}, such as VideoPhy~\citep{bansal2024videophyevaluatingphysicalcommonsense, bansal2025videophy2challengingactioncentricphysical} and VideoScore~\citep{he2024videoscorebuildingautomaticmetrics, he2025videoscore2thinkscoregenerative}, to assess physical understanding in multimodal models. These approaches typically formulate evaluation as recognition-based tasks like VQA. While effective for probing high-level physical knowledge, such protocols make it difficult to determine whether model performance reflects genuine physical reasoning or reliance on appearance-based heuristics and dataset-specific biases. Our framework complements previous works by requiring explicit and executable physical hypotheses evaluated through simulation. Table~\ref{tab:mllm-visual-eval} compares our work with prior works.

\textbf{Executable World Representations for Visual and Motion Generation.}
Representing visual scenes as executable programs is a foundational paradigm in computer graphics and simulation, where structured code specifies objects, motion, and physical interactions to enable interpretable and controllable world representations~\citep{foley1996computer}. Recent advances in multimodal large language models have begun to enable the generation of executable code for visual content. Early efforts primarily focus on static visualizations, such as data plots and vector graphics, translating high-level semantic intent into low-level graphical instructions~\citep{galimzyanov2025drawing, yang2024matplotagent, goswami2025plotgen, ni2025viscoderfinetuningllmsexecutable, ni2025viscoder2buildingmultilanguagevisualization, rodriguez2024starvector, yang2025omnisvg, vcode}. These methods demonstrate the feasibility of using code as a structured intermediate between language and visual output. Subsequent work extends code-based generation to animations and motion, enabling programmatic specification of object trajectories and temporal behaviors~\citep{zhang2023motiongraphics, he2024kubrick, liu2024physgen, lv2024gpt4motionscriptingphysicalmotions, ku2025theoremexplainagentmultimodalexplanationsllm}. While these approaches show that MLLM can generate executable programs that produce coherent motion, they are primarily designed for content creation or presentation, and rarely assess whether the generated programs correspond to physically consistent dynamics or reflect an underlying understanding of physical laws. In contrast to prior work that treats executable visual generation as an end goal, our work uses executable world representations as a diagnostic interface for physical reasoning. Rather than evaluating visual realism or animation quality, we assess whether models can reconstruct and resimulate physically consistent dynamics from visual observations to enable direct inspection.

\begin{table}[!htbp]
\centering
\caption{
VisPhyWorld uniquely turns physical reasoning into an executable hypothesis and enables multimetric, diagnostic evaluation beyond relative scoring.
Gen.: requires future visual generation; MLLM: evaluates MLLM outputs;
Exec.: requires an executable physical hypothesis.
}
\label{tab:mllm-visual-eval}
\footnotesize
\setlength{\tabcolsep}{3.5pt}
\renewcommand{\arraystretch}{1.08}

\begin{tabular}{@{}lccccc@{}}
\toprule
Benchmark & Gen. & MLLM & Exec. & Scoring & Output \\
\midrule
PHYRE~\citep{bakhtin2019phyre}
& \cross & \cross & \cross & Relative & Action \\
CLEVRER~\citep{yi2020clevrercollisioneventsvideo}
& \tick & \cross & \cross & Relative & QA / VoE \\
IntPhys~\citep{riochet2018intphys}
& \tick & \cross & \cross & Relative & VoE \\
PhyGenBench~\citep{meng2024worldsimulatorcraftingphysical}
& \tick & \tick & \cross & Relative & QA \\
MVP~\citep{krojer2025shortcutawarevideoqabenchmarkphysical}
& \cross & \tick & \cross & Relative & QA \\
PhysicsIQ~\citep{motamed2025generativevideomodelsunderstand_physicsiq}
& \tick & \tick & \cross & Relative & QA \\
WorldModelBench~\citep{li2026worldmodelbench}
& \tick & \tick & \cross & VLM judge & Video \\
IntPhys2~\citep{bordes2025intphys}
& \cross & \tick & \cross & Relative & VoE \\
PhyWorld~\citep{kang2024far}
& \tick & \cross & \cross & Reconstruction & Video \\
\midrule
\rowcolor{blue!8}
\textbf{VisPhyWorld (Ours)}
& \tick & \tick & \tick & \textbf{Reconstruction} & \textbf{Code} \\
\bottomrule
\end{tabular}
\end{table}

\section{VisPhyWorld}
\label{sec:VisPhyWorld}

We introduce VisPhyWorld, a framework that uses MLLM to interpret visual observations and reconstruct the underlying physical scene as executable code.
We evaluate the rendered outputs under a unified protocol using a multi-metric suite.

\begin{figure}[t]
    \centering
    \ifvisTikz
\begin{minipage}{\columnwidth}
\centering
\begin{minipage}[c]{0.56\columnwidth}
\centering
\tiny
\setlength{\tabcolsep}{2.0pt}
\renewcommand{\arraystretch}{1.15}
\resizebox{\linewidth}{!}{%
\begin{tabular}{lcccccc}
\toprule
\textbf{Engine} &
\makecell{\textbf{LPIPS}\\$\downarrow$} &
\makecell{\textbf{CLIP}\\\textbf{Img}$\uparrow$} &
\makecell{\textbf{DINO}\\$\uparrow$} &
\makecell{\textbf{CLIP}\\\textbf{Cap}$\uparrow$} &
\makecell{\textbf{BERT}\\\textbf{Score}$\uparrow$} &
\makecell{\textbf{RAFT}\\\textbf{EPE}$\downarrow$} \\
\midrule
Three.js & 0.16 & 0.89 & 0.86 & 0.27 & 0.84 & 31.08 \\
SVG      & 0.16 & 0.92 & 0.87 & 0.26 & 0.85 & 30.06 \\
Blender  & 0.25 & 0.74 & 0.71 & 0.31 & 0.83 & 30.18 \\
P5.js    & 0.24 & 0.84 & 0.79 & 0.23 & 0.81 & 36.71 \\
Manim    & 0.28 & 0.85 & 0.78 & 0.25 & 0.84 & 31.73 \\
\bottomrule
\end{tabular}%
}
\end{minipage}
\hfill
\begin{minipage}[c]{0.42\columnwidth}
\centering
\begin{minipage}[c]{0.72\linewidth}
\centering
\resizebox{\linewidth}{!}{%
\begin{tikzpicture}[font=\scriptsize]
  \def\R{1.9500}
  \draw[gray!28] (0,0) circle ({\R*0.2});
  \draw[gray!28] (0,0) circle ({\R*0.4});
  \draw[gray!28] (0,0) circle ({\R*0.6});
  \draw[gray!28] (0,0) circle ({\R*0.8});
  \draw[gray!28] (0,0) circle ({\R*1.0});
  \draw[gray!45] (0,0) -- (90.0000:\R);
  \node[align=center] at (90.0000:{\R+0.34}) {LPIPS $\downarrow$};
  \draw[gray!45] (0,0) -- (38.5714:\R);
  \node[align=center] at (38.5714:{\R+0.34}) {CLIP-Img $\uparrow$};
  \draw[gray!45] (0,0) -- (-12.8571:\R);
  \node[align=center] at (-12.8571:{\R+0.34}) {DINO $\uparrow$};
  \draw[gray!45] (0,0) -- (-64.2857:\R);
  \node[align=center] at (-64.2857:{\R+0.34}) {CLIP-Cap $\uparrow$};
  \draw[gray!45] (0,0) -- (-115.7143:\R);
  \node[align=center] at (-115.7143:{\R+0.34}) {BERTScore-F1 $\uparrow$};
  \draw[gray!45] (0,0) -- (-167.1429:\R);
  \node[align=center] at (-167.1429:{\R+0.34}) {RAFT-EPE $\downarrow$};
  \draw[gray!45] (0,0) -- (-218.5714:\R);
  \node[align=center] at (-218.5714:{\R+0.34}) {Gemini $\uparrow$};
  \draw[blue!75!black, line width=0.75pt] (90.0000:1.6883) -- (38.5714:1.9208) -- (-12.8571:1.7590) -- (-64.2857:1.4335) -- (-115.7143:0.3900) -- (-167.1429:1.8358) -- (-218.5714:1.7525) -- (90.0000:1.6883);
  \fill[blue!75!black, opacity=0.06] (90.0000:1.6883) -- (38.5714:1.9208) -- (-12.8571:1.7590) -- (-64.2857:1.4335) -- (-115.7143:0.3900) -- (-167.1429:1.8358) -- (-218.5714:1.7525) -- (90.0000:1.6883);
  \draw[cyan!65!black, line width=0.75pt] (90.0000:1.6246) -- (38.5714:1.9228) -- (-12.8571:1.5889) -- (-64.2857:1.2016) -- (-115.7143:1.9500) -- (-167.1429:1.8280) -- (-218.5714:1.4629) -- (90.0000:1.6246);
  \fill[cyan!65!black, opacity=0.06] (90.0000:1.6246) -- (38.5714:1.9228) -- (-12.8571:1.5889) -- (-64.2857:1.2016) -- (-115.7143:1.9500) -- (-167.1429:1.8280) -- (-218.5714:1.4629) -- (90.0000:1.6246);
  \draw[green!60!black, line width=0.75pt] (90.0000:1.9500) -- (38.5714:1.9500) -- (-12.8571:1.6570) -- (-64.2857:0.7484) -- (-115.7143:0.8253) -- (-167.1429:1.5230) -- (-218.5714:1.9500) -- (90.0000:1.9500);
  \fill[green!60!black, opacity=0.06] (90.0000:1.9500) -- (38.5714:1.9500) -- (-12.8571:1.6570) -- (-64.2857:0.7484) -- (-115.7143:0.8253) -- (-167.1429:1.5230) -- (-218.5714:1.9500) -- (90.0000:1.9500);
  \draw[orange!75!black, line width=0.75pt] (90.0000:1.7924) -- (38.5714:1.9391) -- (-12.8571:1.5895) -- (-64.2857:0.9697) -- (-115.7143:0.9705) -- (-167.1429:1.5236) -- (-218.5714:1.0219) -- (90.0000:1.7924);
  \fill[orange!75!black, opacity=0.06] (90.0000:1.7924) -- (38.5714:1.9391) -- (-12.8571:1.5895) -- (-64.2857:0.9697) -- (-115.7143:0.9705) -- (-167.1429:1.5236) -- (-218.5714:1.0219) -- (90.0000:1.7924);
  \draw[purple!70!black, line width=0.75pt] (90.0000:1.3226) -- (38.5714:1.7761) -- (-12.8571:1.2736) -- (-64.2857:1.6232) -- (-115.7143:0.9342) -- (-167.1429:1.6642) -- (-218.5714:0.8442) -- (90.0000:1.3226);
  \fill[purple!70!black, opacity=0.06] (90.0000:1.3226) -- (38.5714:1.7761) -- (-12.8571:1.2736) -- (-64.2857:1.6232) -- (-115.7143:0.9342) -- (-167.1429:1.6642) -- (-218.5714:0.8442) -- (90.0000:1.3226);
  \draw[red!70!black, line width=0.75pt] (90.0000:0.3900) -- (38.5714:0.3900) -- (-12.8571:0.3900) -- (-64.2857:0.3900) -- (-115.7143:0.3900) -- (-167.1429:0.3900) -- (-218.5714:0.3900) -- (90.0000:0.3900);
  \fill[red!70!black, opacity=0.06] (90.0000:0.3900) -- (38.5714:0.3900) -- (-12.8571:0.3900) -- (-64.2857:0.3900) -- (-115.7143:0.3900) -- (-167.1429:0.3900) -- (-218.5714:0.3900) -- (90.0000:0.3900);
  \draw[gray!75!black, line width=0.75pt] (90.0000:1.4041) -- (38.5714:1.6721) -- (-12.8571:1.9500) -- (-64.2857:1.9500) -- (-115.7143:0.3900) -- (-167.1429:1.9500) -- (-218.5714:1.1733) -- (90.0000:1.4041);
  \fill[gray!75!black, opacity=0.06] (90.0000:1.4041) -- (38.5714:1.6721) -- (-12.8571:1.9500) -- (-64.2857:1.9500) -- (-115.7143:0.3900) -- (-167.1429:1.9500) -- (-218.5714:1.1733) -- (90.0000:1.4041);
\end{tikzpicture}%
}
\end{minipage}
\hfill
\begin{minipage}[c]{0.25\linewidth}
\centering
\tiny
\setlength{\tabcolsep}{1.1pt}
\renewcommand{\arraystretch}{0.95}
\begin{tabular}{@{}ll@{}}
{\color{blue!75!black}\rule{0.8em}{0.30mm}} & GPT-5 \\
{\color{cyan!65!black}\rule{0.8em}{0.30mm}} & GPT-4.1 \\
{\color{green!60!black}\rule{0.8em}{0.30mm}} & Gemini \\
{\color{orange!75!black}\rule{0.8em}{0.30mm}} & Claude \\
{\color{purple!70!black}\rule{0.8em}{0.30mm}} & Qwen3 \\
{\color{red!70!black}\rule{0.8em}{0.30mm}} & SVD \\
{\color{gray!75!black}\rule{0.8em}{0.30mm}} & Veo \\
\end{tabular}
\end{minipage}
\end{minipage}
\end{minipage}
    \else
      \fbox{\parbox{0.95\linewidth}{\centering TikZ figure omitted (missing tikz/pgf in this TeX installation).}}
    \fi
    \caption{
    Left: cross-engine GPT-5 stress-test scores. Right: radar summary of key VisPhyBench metrics comparing code-driven reconstruction against pixel-space baselines. Engine choice changes absolute visual scores, but the scene-level motion difficulty ranking is stable across engines. Therefore, our conclusions are not merely driven by backend-specific coding artifacts.}
    \label{fig:visphybench-radar}
\end{figure}

\subsection{Problem Definition}
\label{sec:problem}

We focus on 2D and 3D physical scenes involving common interactions, e.g., ball collisions and box sliding.
We represent each scene as a sequence of image frames $I$ with three color channels as in Equation~\ref{eq:rgb_video}, where $H$, $W$, and $T$ denote frame height, frame width, and number of frames.
\begin{equation}
    X = (I_t)_{t=1}^{T}, \qquad I_t \in \mathbb{R}^{3 \times H \times W},
    \label{eq:rgb_video}
\end{equation}

\textbf{Input.} Given a scene, the MLLM backbone receives $\{I^{\text{start}}, I^{\text{later}}, D\}$.
We select two key frames from $X$, where $I^{\text{start}} = I_{t_s}$ and $I^{\text{later}} = I_{t_l}$, typically corresponding to an early frame and a later frame (e.g., $t_s = 1$, $t_l = 10$).
Optionally, we provide a detection context $D$ for $I^{\text{start}}$ listing objects with categories, bounding boxes, and coarse attributes (see Appendix~\ref{app:detection-context}).
We obtain $D$ with GPT-5.2~\citep{openai_gpt52_model} on the first frame and parsing its output into a structured object list; if unavailable, we set $D = \varnothing$.

\textbf{Outputs.}
VisPhyWorld produces four interpretable artifacts:
(i) a textual motion analysis $A \in \mathcal{Y}^{\text{text}}$;
(ii) a machine-readable first-frame JSON specification $S$ encoding object layout and inferred parameters;
(iii) an executable program $C \in \mathcal{Y}^{\text{code}}$; and
(iv) a rendered video $\hat{X} = (\hat{I}_t)_{t=1}^{\hat{T}}$ obtained by executing the executable program $C$.

\subsection{VisPhyWorld Architecture}
\label{sec:architecture}
\begin{equation}
    (I^{\text{start}}, I^{\text{later}}, D)
    \xrightarrow{f_{\text{LLM}}}
    (A, S, C)
    \xrightarrow{R_{\text{phys}}}
    \hat{X}.
    \label{eq:mapping}
\end{equation}

VisPhyWorld implements a composite mapping as stated in Equation~\ref{eq:mapping}. We include $A$ as a lightweight, text-only diagnostic of the model's basic scene understanding: whether it can correctly describe salient motions and interactions between the key frames, separately from code generation.
We treat $C$ as an explicit, falsifiable physical hypothesis: executing it with a renderer $R_{\text{phys}}$ under a fixed configuration yields $\hat{X}$, separating hypothesis construction from execution and enabling controlled comparisons across LLM backbones.
To ensure a well-defined evaluation, we apply lightweight validation prior to execution and allow a single automatic repair attempt upon failure; if both attempts fail, we fall back to a minimal valid scene. Further implementation details, including prompt templates and renderer settings, are deferred to Appendix~\ref{app:implementation}, with robustness analyses in Appendix~\ref{app:robustness}.

\subsection{Benchmark, Metrics, and Baselines}
\label{sec:benchmark}

\noindent\textbf{Dataset Construction.}
We build on and enrich the 2D data from the PhyWorld dataset~\citep{kang2024far}, using the PHYRE engine~\citep{bakhtin2019phyre} for rendering to form the 2D subset of VisPhyBench. We additionally curate a 3D subset rendered with Three.js and simulated using Cannon.js for rigid-body dynamics. Overall, VisPhyBench comprises 108 templates and 209 videos, each paired with first-frame JSON annotations.
VisPhyBench scenes are annotated with coarse difficulty levels. We construct a small test split by subsampling from the full dataset to enable rapid sanity checks and lightweight evaluation. Eight STEM-trained annotators rate each raw clip on a 1–5 scale (higher indicates greater difficulty), and we use the mean rating as the final difficulty score. The mean score is then mapped to easy, medium, or hard using fixed, interpretable cutoffs aligned with the rating scale (easy = 1–2, medium = 3, hard = 4–5). The resulting distribution is naturally skewed, reflecting the relative rarity of challenging interactions in our template set: the \texttt{sub} split contains 114 easy, 67 medium, and 28 hard scenes, while the \texttt{test} split contains 29 easy, 17 medium, and 3 hard scenes. Scenes cover diverse object configurations (stacks, ramps, collisions) and motion patterns (slides, bounces, topples).
For the 2D subset, the camera is fixed and orthographic; for the 3D subset, we use a fixed perspective camera.
In both settings, the background is set to white to focus on physical dynamics. Since templates are executable programs instantiated by sampling seeds, we summarize template composition and object statistics in Appendix~\ref{app:visphybench-templates}.
Inputs $(I^{\text{start}}, I^{\text{later}}, D)$ follow Section~\ref{sec:problem}.

\noindent\textbf{Evaluation Metrics.}
We report per-metric means over all scenes and group metrics into five families.
\begin{enumerate*}[label=\textbf{(\arabic*)}, itemjoin={{\quad}}, itemjoin*={{\quad}}]
\item Reconstruction and perceptual quality.
We report PSNR~\citep{article} and SSIM~\citep{1284395} for frame-wise reconstruction, together with LPIPS~\citep{zhang2018unreasonableeffectivenessdeepfeatures}, FSIM~\citep{5705575}, VSI~\citep{6873260}, and DISTS~\citep{Ding_2020} to compute on aligned frame pairs.
\item Visual semantic consistency.
We compute CLIP-based image similarity (CLIP-Img)~\citep{radford2021learningtransferablevisualmodels} and DINO feature similarity~\citep{caron2021emergingpropertiesselfsupervisedvision}, which emphasize object identity and scene layout beyond exact pixels.
\item Text--video and analysis-text consistency.
We compute CLIP text--image similarity (CLIP-Cap)~\citep{radford2021learningtransferablevisualmodels} between the analysis and sampled video frames, and use ROUGE-L~\citep{lin-2004-rouge} and BERTScore-F1~\citep{zhang2020bertscoreevaluatingtextgeneration} to compare the analysis with a GPT-generated reference description derived from the ground-truth video.
\item Motion and physical plausibility.
We use RAFT-based optical-flow diagnostics~\citep{teed2020raftrecurrentallpairsfield} with automatic temporal alignment, reporting RAFT end-point error (EPE) and the estimated temporal offset (and, when relevant, flow magnitude and angular statistics) to quantify motion consistency.
Because flow discrepancy alone can be misleading as a proxy for physical plausibility, we interpret RAFT metrics jointly with holistic perceptual/physics judgments rather than using RAFT-EPE in isolation; as discussed in Section~\ref{sec:eval-metrics}.
\item Subjective overall quality.
We use a Gemini-2.5-Pro video--video judge (1--10) with a textual justification, and report a model-blind human evaluation score (1--5); details and calibration against other judges are provided in Appendix~\ref{sec:app-gemini}. We separately report pipeline success rate based on whether a valid video is produced.
\end{enumerate*}

\noindent\textbf{Video Model Baselines.}
We include Stable Video Diffusion (SVD) img2vid~\citep{blattmann2023stablevideodiffusionscaling}, conditioned only on $I^{\text{start}}$, Sora-2, Veo-3.1, and Cosmos-Predict2.5-2B~\citep{nvidia2025cosmosworldfoundationmodel}, a recent world-foundation model targeted at physical understanding, with the direct video models conditioned on the same visual frames and prompts allowed by their APIs.

\subsection{Engine Evaluation and Selection}
\label{sec:ablation-engines}

We evaluate four rendering backends, \ie, Three.js~\citep{threejs}, P5.js~\citep{p5js}, SVG (Scalable Vector Graphics), and Manim~\citep{The_Manim_Community_Developers_Manim_Mathematical_2024}, to understand how the choice of visualization engine affects multimodal LLM-based reconstruction.
As shown in Figure~\ref{fig:engine_eval_task10010}, a consistent pattern emerges: Three.js and P5.js achieve markedly better reconstruction and motion fidelity because they support native integration with rigid-body physics solvers, allowing the generated programs to offload gravity, contact constraints, friction, and collision response to a physically grounded engine.
In contrast, SVG and Manim are primarily non-physics-based rendering systems: they excel at deterministic drawing and scripted animation, but lack intrinsic rigid-body dynamics. In our experimental setting, SVG and Manim serve as non-interactive, script-based backends and do not expose a comparable physics API or closed-loop simulation stepping; consequently, as illustrated in Figure~\ref{fig:engine_eval_task10010}, they often yield physically implausible behaviors, such as objects remaining static or interpenetrating.
Importantly, this gap suggests a limitation of current MLLM: without access to a true physics solver, they fail to consistently infer and apply Newtonian dynamics from visual evidence, and instead revert to heuristic motion scripting.
For this work, we therefore prioritize Three.js and P5.js so that our evaluation emphasizes physically grounded re-simulation rather than non-physical animation artifacts.
To check that the conclusions are not merely backend-specific coding artifacts, we run a cross-engine stress test with GPT-5 across Three.js, P5.js, SVG, Manim, and Blender. The results are shown in the left panel of Figure~\ref{fig:visphybench-radar}, with correlation matrices reported in Appendix~\ref{app:cross-engine-robustness}. Scene-level RAFT-EPE rankings are highly correlated across engines, with a mean Spearman correlation of $\rho=0.84$ and all pairwise $p<0.05$, while appearance metrics are only moderately correlated. These results suggest that our conclusions are not solely driven by backend-specific coding artifacts.

\section{Experiments}
\label{sec:experiments}

\noindent\textbf{Evaluation setup.}
We evaluate VisPhyWorld and all baselines on VisPhyBench. Local rendering, post-processing, and metric evaluation were run on a workstation with 4 NVIDIA RTX A6000 GPUs. API-based models were queried through their provider-hosted services.
For each configuration, we generate one video per scene, compute all metrics, and report per-metric means over the evaluation split; unless otherwise stated, higher is better.
We consider five multimodal LLM backbones: GPT-5~\citep{openai_gpt5}, GPT-4.1~\citep{openai_gpt41}, Gemini-3-Pro~\citep{google_gemini3}, Claude Sonnet 4.5~\citep{anthropic_claude_sonnet45}, and Qwen3-VL-Plus~\citep{alibaba_qwen3_vl_plus}.
We evaluate two code backends, Three.js~\citep{threejs} and P5.js~\citep{p5js}. All LLM runs use the same prompt and two key frames; only the model and
engine identifiers change. For each run, we aggregate metrics into five families: reconstruction \& perceptual quality,
visual semantic consistency, text--video \& analysis-text consistency, motion (automatic RAFT-based metrics~\citep{teed2020raftrecurrentallpairsfield}), and subjective
overall quality (Gemini-2.5-Pro~\citep{comanici2025gemini25pushingfrontier} judge). We observe consistent trends across scenes; per-scene metric distributions and significance analyses are reported in Appendix~\ref{app:distributions-significance}, Fig.~\ref{fig:visphybench-metric-distributions}.

\subsection{Overall leaderboard.}
Table~\ref{tab:visphybench-leaderboard} summarizes performance across five metric families on VisPhyBench.
Overall, most models achieve strong reconstruction and perceptual scores and maintain reasonable visual-semantic consistency; these results support our central claim that, once the task is cast as executable hypotheses under a fixed physics engine, most modern MLLM can reconstruct synthetic physical events with high fidelity, and the remaining gaps become diagnosable rather than opaque.

\noindent\textbf{Visual and language-based reasoning can diverge.}
Our benchmark jointly evaluates visual reconstruction, visual semantics, and language-mediated reasoning, and we find that these dimensions do not always move together.
Some model--backend pairs achieve strong perceptual and semantic alignment with the reference frames, as shown by low LPIPS and high CLIP-Img and DINO scores.
These results suggest that they can recover object identities and global layouts well from the visual input.
For example, Gemini-3-Pro with Three.js achieves the lowest LPIPS and the highest CLIP-Img score, and it also obtains the strongest pixel-level reconstruction in Appendix Table~\ref{tab:app-psnr-ssim-success}.
However, strong visual reconstruction does not necessarily imply equally strong language-based reasoning.
For instance, GPT-4.1 with Three.js achieves the highest BERT-F1 score, even though its LPIPS is higher than that of Gemini-3-Pro with Three.js.
This gap shows that VisPhyBench is not only measuring overall model quality.
Instead, it separates a model's ability to see the scene from its ability to explain the scene.
In contrast, video-generation baselines such as Veo-3.1 do not produce executable simulators, which makes their intermediate states less transparent and harder to diagnose.

\begin{table}[t]
    \caption{
        \textbf{Overall leaderboard on VisPhyBench}.
        $^{\mathrm{T}}$ denotes Three.js and $^{\mathrm{P}}$ denotes p5.js.
        Higher is better ($\uparrow$), lower is better ($\downarrow$).
        ``--'' denotes unavailable or inapplicable metrics.
    }
    \centering
    \footnotesize
    \setlength{\tabcolsep}{3pt}
    \renewcommand{\arraystretch}{1.08}

    \begin{adjustbox}{max width=\textwidth,center}
    \begin{tabular}{@{} >{\raggedright\arraybackslash}p{0.18\textwidth} cccccccc @{}}
        \toprule
        \multirow{2}{*}{\textbf{Model}} &
        \multicolumn{1}{c}{\makecell{\textbf{Reconst.}\\\textbf{\& Perceptual}}} &
        \multicolumn{2}{c}{\makecell{\textbf{Visual Semantic}\\\textbf{Consistency}}} &
        \multicolumn{2}{c}{\makecell{\textbf{Text--Video}\\\textbf{\& Analysis-Text}}} &
        \multicolumn{1}{c}{\makecell{\textbf{Motion / Physical}\\\textbf{Plausibility}}} &
        \multicolumn{1}{c}{\makecell{\textbf{Holistic}\\\textbf{Quality}}} &
        \multicolumn{1}{c}{\makecell{\textbf{Human}\\\textbf{Preference}}} \\
        \cmidrule(lr){2-2}
        \cmidrule(lr){3-4}
        \cmidrule(lr){5-6}
        \cmidrule(lr){7-7}
        \cmidrule(lr){8-8}
        \cmidrule(lr){9-9}
        & \textbf{LPIPS$\downarrow$} &
          \textbf{CLIP-Img$\uparrow$} &
          \textbf{DINO$\uparrow$} &
          \textbf{CLIP-Cap$\uparrow$} &
          \textbf{BERT-F1$\uparrow$} &
          \textbf{RAFT-EPE$\downarrow$} &
          \textbf{Gemini$\uparrow$} &
          \textbf{Human$\uparrow$} \\
        \midrule
        GPT-5$^{\mathrm{T}}$             & 0.17 & 0.89 & 0.86          & 0.26          & 0.84 & 33.65 & 3.50 & 4.33 \\
        GPT-5$^{\mathrm{P}}$             & 0.29 & 0.81 & 0.76          & 0.23          & 0.84 & 34.34 & 3.52 & 3.67 \\
        GPT-4.1$^{\mathrm{T}}$           & 0.18 & 0.89 & 0.83          & 0.26          & \textbf{0.85} & 33.71 & 3.06 & 4.17 \\
        GPT-4.1$^{\mathrm{P}}$           & 0.35 & 0.75 & 0.68          & 0.22          & 0.83 & 37.70 & 2.15 & 2.33 \\
        Gemini-3-Pro$^{\mathrm{T}}$      & \textbf{0.14} & \textbf{0.90} & 0.84 & 0.26 & 0.85 & 36.20 & \textbf{3.80} & \textbf{4.67} \\
        Gemini-3-Pro$^{\mathrm{P}}$      & 0.33 & 0.75 & 0.67          & 0.22          & 0.84 & 33.10 & 2.35 & 3.50 \\
        Claude-S4.5$^{\mathrm{T}}$       & 0.16 & 0.90 & 0.83          & 0.26          & 0.85 & 36.20 & 2.39 & 3.83 \\
        Claude-S4.5$^{\mathrm{P}}$       & 0.33 & 0.76 & 0.71          & 0.22          & 0.82 & 34.14 & 2.56 & 2.67 \\
        Qwen3-VL-Plus$^{\mathrm{T}}$     & 0.22 & 0.87 & 0.78          & \textbf{0.27} & 0.85 & 35.05 & 2.12 & 2.50 \\
        Qwen3-VL-Plus$^{\mathrm{P}}$     & 0.55 & 0.64 & 0.55          & 0.20          & 0.84 & \textbf{20.82} & 1.46 & 1.83 \\
        \midrule
        SVD (img2vid)                    & 0.34 & 0.67 & 0.65          & 0.25          & --   & 45.46 & 1.43 & 1.83 \\
        Cosmos-2.5-2B             & 0.27 & 0.73 & 0.65          & --            & --   & 33.12 & 1.25 & 1.16 \\
        Sora-2                           & 0.20 & 0.87 & 0.87          & 0.26          & --   & 34.91 & 2.35 & 2.83 \\
        Veo-3.1                          & 0.21 & 0.86 & \textbf{0.88} & \textbf{0.27} & --   & 32.71 & 2.62 & 4.50 \\
        \bottomrule
    \end{tabular}
    \end{adjustbox}

    \label{tab:visphybench-leaderboard}
\end{table}

\noindent\textbf{The code backend affects reconstruction quality.}
Across LLMs, Three.js variants generally reconstruct the scenes better than their p5.js counterparts, as shown by lower LPIPS scores in most model pairs, even though they use the same inputs and prompts.
For example, for GPT-5, using Three.js reduces LPIPS by nearly 40\% and improves SSIM from $0.74$ to $0.94$.
Qualitatively, this leads to more stable object identities and cleaner scene layouts, as shown in Figure~\ref{fig:engine_eval_task10010}.
Since the physical setup is kept the same, this gap suggests that the backend interface and program structure affect how well a model can translate visual evidence into an executable simulation.
In other words, the benchmark captures not only physical reasoning, but also whether the chosen simulator provides a clear and controllable way for the model to express that reasoning.

\noindent\textbf{Pixel-space baselines show a complementary but less interpretable profile.} They can score competitively on some feature semantics, but their failures are harder to attribute to specific physical causes, such as friction, restitution, or contact timing, since the generation process does not expose interpretable latent variables.
Veo-3.1 attains reasonable semantic similarity, for example reaching DINO $\sim0.88$, yet it does not expose an explicit simulator state for diagnosis or controlled interventions and often exhibits deficiencies in physical understanding by producing trajectories with implausible motion or contact events (see Sec.~\ref{sec:qualitative-case-study}).
This pattern also holds for world-foundation models designed for physical understanding. Cosmos-Predict2.5-2B has the lowest visual scores and the worst Gemini and human ratings among all baselines. Qualitatively, it often fails to keep object identities consistent across frames. This shows that even physics-aware video generators can still struggle with stable physical rollout, while code-based simulators remain more transparent and controllable.
Conversely, our code-driven approach maintains competitive semantic and motion scores while exposing executable states; e.g., GPT-5 (threejs) achieves DINO $0.8556$ with RAFT-EPE $33.6473$.
This enables controlled interventions (e.g., varying friction/mass while holding layout fixed) that can isolate whether an error originates from object discovery, state initialization, or contact modeling, aligning with our goal of turning ``physics understanding'' into a testable, executable object.

\noindent\textbf{Motion and holistic physical plausibility.}
\label{sec:eval-metrics}
Assessing physical plausibility requires combining motion-level metrics with holistic perceptual judgment.
We report RAFT-EPE to measure optical-flow discrepancy, as detailed in Appendix~\ref{app:metrics-details}, and use a Gemini-2.5-Pro judge to aggregate visual and physical cues into a single holistic score, as detailed in Appendix~\ref{sec:app-gemini}.
RAFT-EPE is useful for measuring trajectory agreement, but relying on it alone can be misleading.
For example, Qwen3-VL-Plus with p5.js obtains the lowest RAFT-EPE, $20.82$, despite weak reconstruction fidelity in Appendix Table~\ref{tab:app-psnr-ssim-success} and a low Gemini score of $1.46$.
In contrast, Gemini-3-Pro with Three.js achieves the highest Gemini score, $3.80$, consistent with its strong visual alignment, although scores around $3$--$4$ still indicate partial scene recovery with remaining motion or contact errors.
Therefore, we treat strong physical understanding as requiring both favorable motion agreement and perceptually coherent outcomes.
The Gemini judge further provides textual justifications that comment on collisions, contact consistency, missing motion, and implausible dynamics, offering a qualitative sanity check alongside the quantitative flow metric.
Together, these metrics provide a multi-view diagnostic evaluation of visual reconstruction, language-based reasoning, and physical plausibility under executable simulation.

\subsection{Robustness and Ablation Studies}
\label{sec:ablation}

\noindent\textbf{Detection context ablation.}
We remove the structured object list $D$ while keeping the same two input frames and prompt style.
As reported in Appendix~\ref{app:additional-ablations}, removing the detection context causes a clear drop in LPIPS and Gemini score across all models, while the other metrics change much less.
This suggests that detection context mainly helps object discovery and state initialization, rather than directly solving physical reasoning.
We therefore include detection context to reduce the confound from object localization.

\noindent\textbf{Judge replacement.}
We also test whether our conclusions depend on the particular evaluator.
VideoScore2~\citep{he2025videoscore2thinkscoregenerative} provides an additional non-LLM evaluation protocol, while Appendix~\ref{app:additional-ablations} reports GPT-5.4 and Qwen3-VL-Plus as alternative holistic judges.
Although the absolute score scales differ across judges, the main ordering remains stable and weak video baselines remain weak.
This supports the robustness of our conclusions to judge replacement.

\noindent\textbf{Forecast-only ablation.}
The forecast-only ablation in Appendix~\ref{app:additional-ablations} removes $I^{\text{later}}$ and uses only $I^{\text{start}}$.
When $I^{\text{later}}$ is removed, visual metrics change little and RAFT-EPE even slightly improves, but the Gemini physical-plausibility score drops more clearly.
This indicates that $I^{\text{later}}$ mainly provides useful physical context.
Without it, models can still generate visually reasonable motion, but are less likely to recover the correct physical outcome.

\begin{table}[htbp]
  \caption{MLLMs often produce visually plausible reconstructions, but fail to recover the physical parameters needed for faithful simulation. Model rows report MAE, with values after $\pm$ denoting standard deviation across scenes.}
  \small
  \centering
  \setlength{\tabcolsep}{5pt}
  \begin{tabular}{lccccc}
    \toprule
    \textbf{Model} &
    \textbf{Gravity} &
    \textbf{Restitution} &
    \textbf{Friction} &
    \makecell{\textbf{Dynamic}\\\textbf{mass}} &
    \makecell{\textbf{Initial}\\\textbf{speed}} \\
    \midrule
    GT mean & 9.82 & 0.15 & 0.40 & 1.25 & 2.59 \\
    GPT-5 & 0 & 0.17$\pm$0.14 & 0.10$\pm$0.07 & 0.15$\pm$0.17 & 2.07$\pm$2.69 \\
    Gemini 3 Pro & 0 & 0.18$\pm$0.13 & 0.11$\pm$0.07 & 0.59$\pm$0.88 & 1.69$\pm$2.42 \\
    Claude Sonnet 4.5 & 0 & 0.16$\pm$0.05 & 0.10$\pm$0.03 & 0.26$\pm$0.42 & 2.39$\pm$3.23 \\
    \bottomrule
  \end{tabular}
  \label{tab:parameter-mae}
\end{table}

\noindent\textbf{Direct parameter recovery.}
Video matching is still an indirect test of physical understanding.
A model may produce a plausible-looking rollout without recovering the latent physical parameters that determine the motion, such as mass, friction, restitution, gravity, and initial velocity.
To measure this more directly, we extract these quantities from the generated simulator code and compare them with the ground-truth parameters using mean absolute error.
Table~\ref{tab:parameter-mae} shows that models recover gravity because it is a stable default, but they make non-trivial errors on contact parameters and much larger errors on initial speed.
Gemini-3-Pro also has a substantial dynamic-mass error despite strong visual reconstruction.
These results show that current MLLMs can often reconstruct the appearance of a physical event, but still fail to infer the quantitative parameters needed for faithful simulation.

\begin{figure}[t]
  \centering
  \begin{minipage}[t]{0.48\textwidth}
    \centering
    \includegraphics[width=\linewidth]{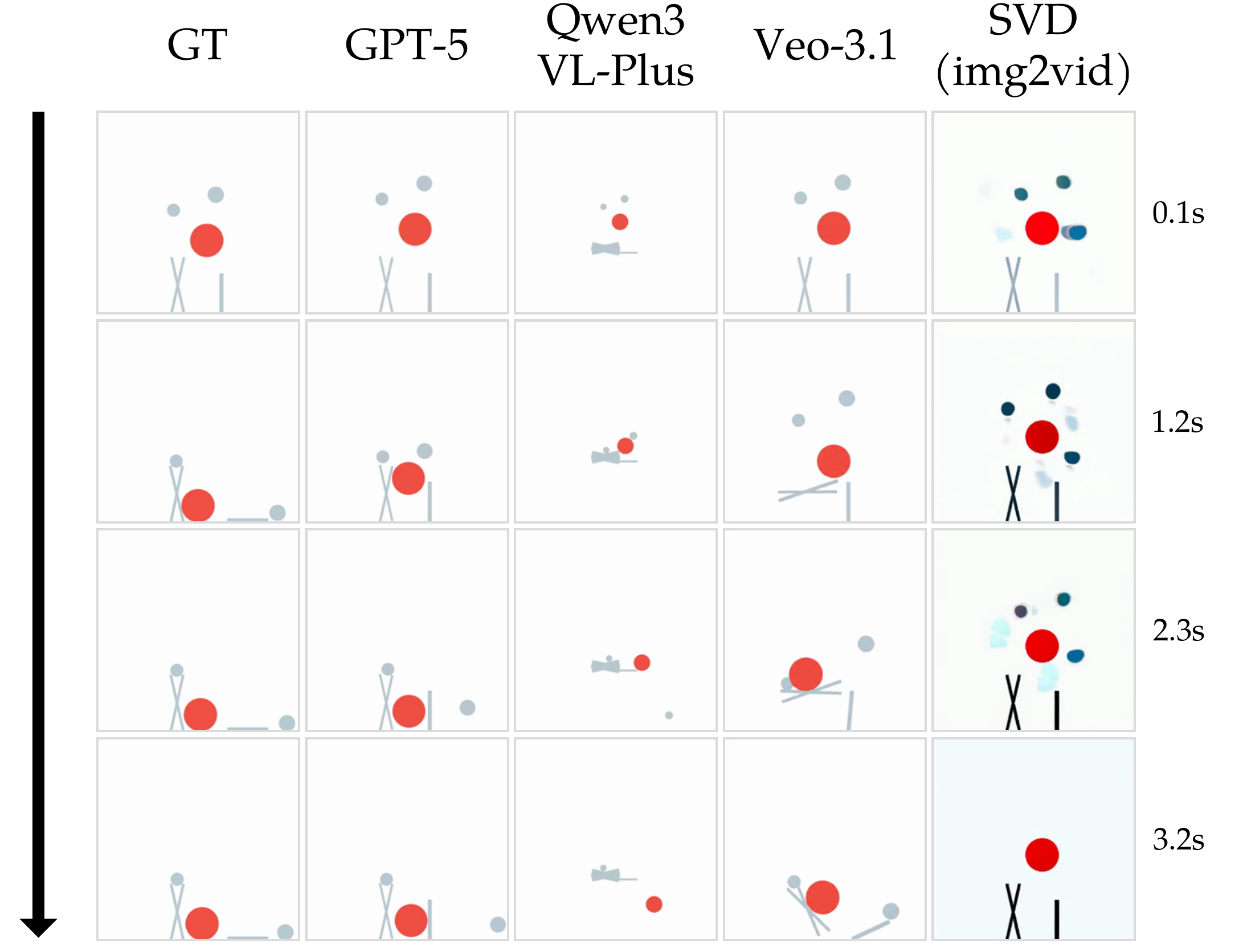}
    \caption{
    GPT-5 reconstructs object identities and collision dynamics most faithfully over time. Pixel-space baselines (Veo-3.1 and SVD/img2vid) generate trajectories with implausible motion/contact events.
    }
    \label{fig:case-study-task10063-000}
  \end{minipage}
  \hfill
  \begin{minipage}[t]{0.48\textwidth}
    \centering
    \includegraphics[width=\linewidth]{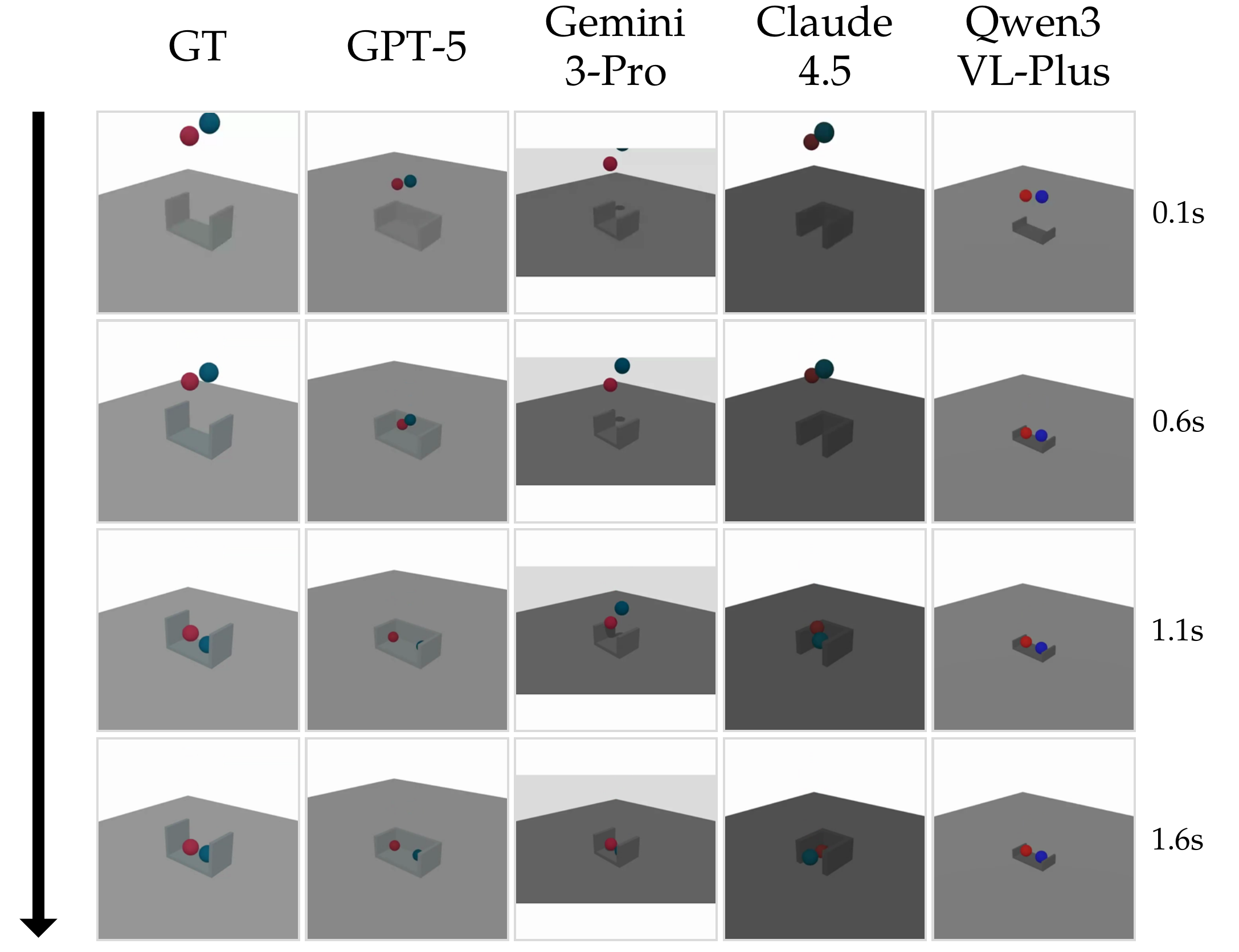}
    \caption{
    This example highlights the dissociation between semantic alignment and correct physical dynamics: although Claude shows clear reconstruction deficits, its visual-semantic scores remain relatively high.
    }
    \label{fig:case-study-dataset3d-14-tmpl-14}
  \end{minipage}
\end{figure}

\subsection{Case Study}
\label{sec:qualitative-case-study}
We present a case study, shown in Fig.~\ref{fig:case-study-task10063-000}, featuring gravity-driven multi-body interactions that require physical reasoning; the diagnostic panel is provided in Appendix Fig.~\ref{fig:case_study}. GPT-5 in Three.js shows strong physical grounding by correctly simulating the collision dynamics, achieving Gemini 10.0 with DINO 0.926. In contrast, pixel-space baselines such as Veo-3.1 achieve high semantic similarity, reaching DINO 0.835, but fail on event logic with Gemini 2.0, indicating plausible appearance with hallucinated dynamics. The case also motivates joint evaluation: Qwen3-VL-Plus attains low RAFT-EPE, 121.30 versus 118.66 for GPT-5, by producing static or empty outputs, yet is penalized by Gemini with a score of 4.0. These results show that optical-flow errors alone are insufficient; credible physical understanding requires both correct motion and holistic visual coherence.

Figure~\ref{fig:case-study-dataset3d-14-tmpl-14} extends our diagnostic analysis beyond 2D templates to a perspective-rendered 3D scene with depth-dependent contacts and occlusions.
Consistent with our 2D findings, we observe the same conclusion in 3D: strong appearance matching alone does not guarantee physically faithful dynamics.
Valid physical understanding is only evidenced when both motion dynamics and holistic visual coherence are satisfied.
For example, Claude-4.5 and Qwen3-VL-Plus exhibit clear reconstruction deviations in this sample, yet their visual-semantic scores do not separate substantially from other models, highlighting a dissociation between semantic alignment and correct physical dynamics.
More broadly, the 3D setting is noticeably more challenging for current MLLMs, underscoring the necessity of incorporating 3D scenes when evaluating reconstruction-based physical reasoning.

\section{Conclusions}
In this work, we introduced VisPhyWorld, a framework that advances the evaluation of physical understanding by requiring MLLMs to reconstruct scenes as executable code, thereby decoupling visual mimicry from physically grounded reasoning. By benchmarking state-of-the-art models on our proposed VisPhyBench, we exposed a consistent dichotomy: while current models excel at semantic scene parsing, they struggle with precise physical parameterization; when forced to commit to an executable hypothesis, models that rely on pixel-space generation often fail to reproduce even basic Newtonian dynamics. Our findings suggest that progress toward robust world modeling may benefit from moving beyond purely statistical pattern matching in pixel space toward hybrid representations that ground visual perception in verifiable, executable physical laws. We believe this direction offers a path toward more transparent and verifiable evaluations of physical understanding.

\clearpage

\bibliographystyle{plainnat}
\bibliography{example_paper}

\clearpage
\appendix


\section*{\Large\bfseries Appendix}
\vspace{5mm}  

\section{Limitations and Discussion}
\label{sec:limitations}

While VisPhyWorld shows promising results on physics-aware video generation and evaluation, it has limitations. First, our experiments are conducted on synthetic, simulator-driven scenes with controlled object layouts and camera motion, so generalization to high-resolution, in-the-wild videos remains untested. Fundamentally limited by the capabilities of current MLLMs and the complexity of modern engines, VisPhyWorld can reliably generate code only for relatively simple rigid-body scenes: although we experimented with large engines such as Unreal Engine, we found that, without human intervention, existing MLLMs cannot, within a small fixed number of calls, autonomously produce and repair simulation code to render a stable, visually plausible video in these more complex environments. Finally, we currently target relatively short clips with moderate motion complexity, and do not address long-horizon interactions, complex 3D reasoning, or stylized or heavily cluttered scenes, which we leave for future work. Future work could integrate stronger 3D perception for scene initialization, and agentic workflows with domain-specific fine-tuning.

\section{Case Study}
\label{app:casestudy}

\begin{figure}[!htbp]
  \centering
  \includegraphics[width=0.7\columnwidth]{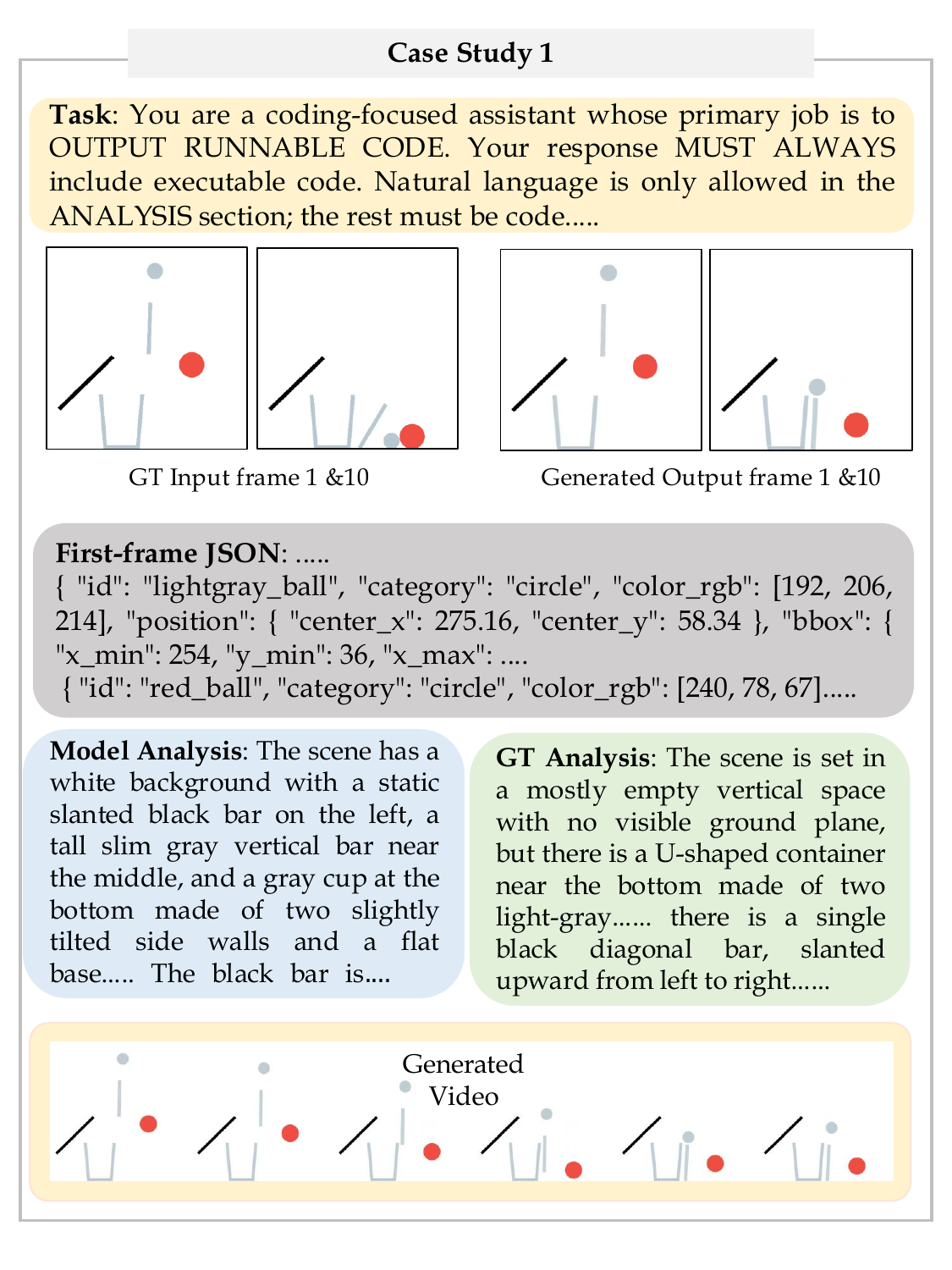}
  \caption{
    Full diagnostic case-study panel. VisPhyWorld exhibits strong physical grounding by reconstructing the scene, intermediate reasoning, generated code, and rendered rollout.
  }
  \label{fig:case_study}
\end{figure}

\clearpage
\begin{figure}[!htbp]
  \centering
  \includegraphics[width=0.95\columnwidth]{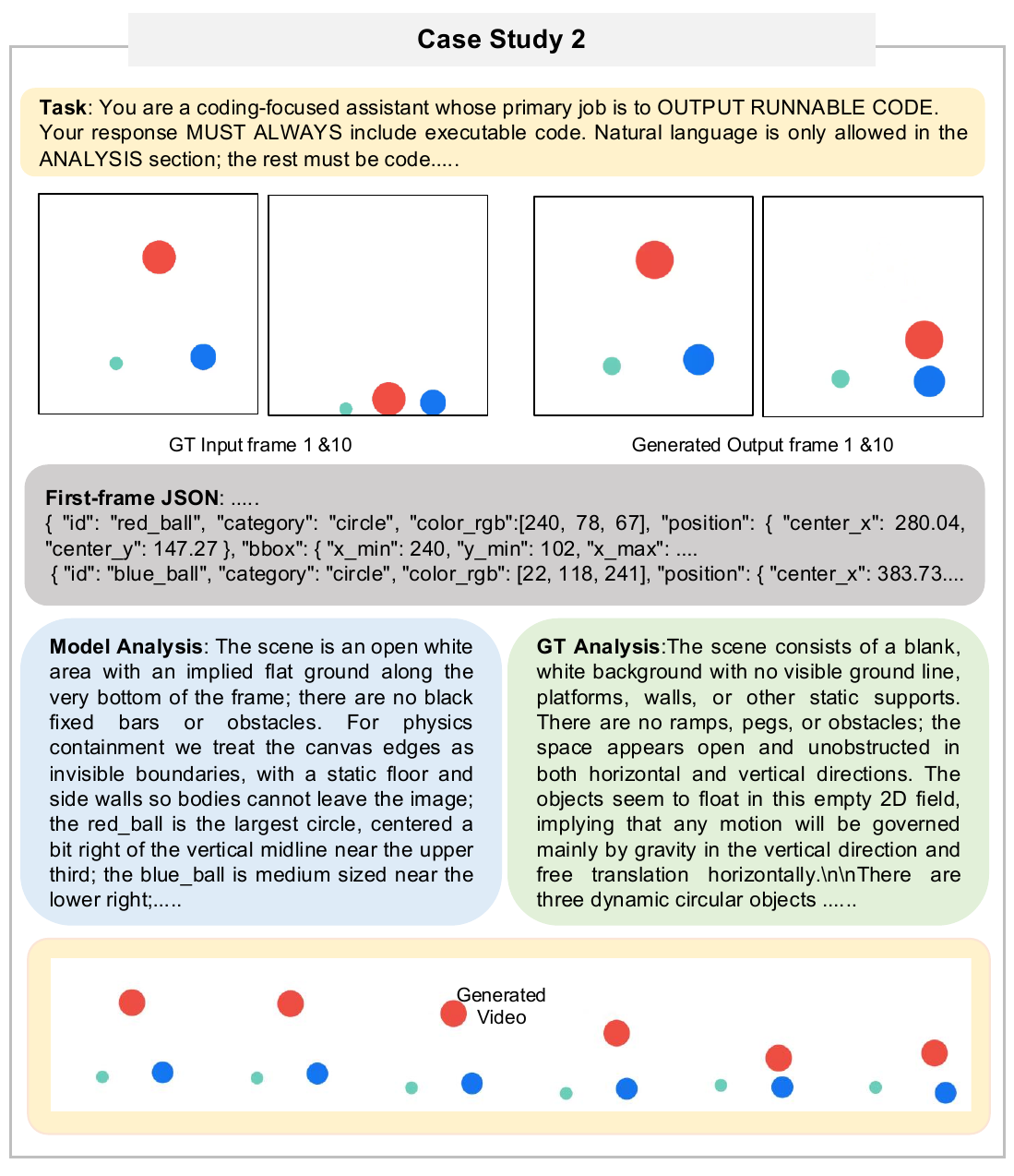}
  \caption{
    A detailed case study (ID 2).
  }
  \label{fig:case_study_2}
\end{figure}

\clearpage
\begin{figure}[t]
  \centering
  \includegraphics[width=0.95\columnwidth]{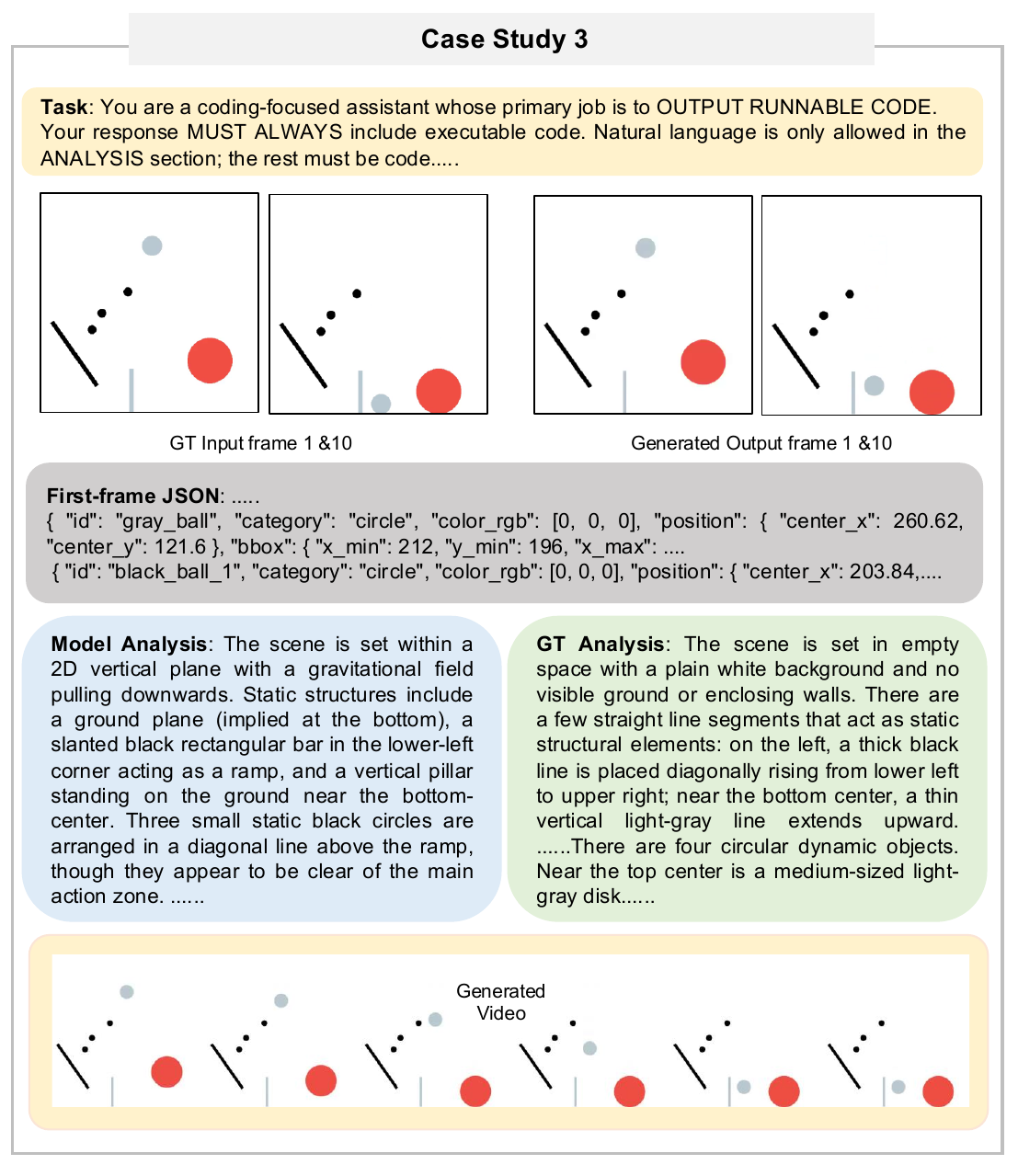}
  \caption{
    A detailed case study (ID 3).
  }
  \label{fig:case_study_3}
\end{figure}
\clearpage

\clearpage
\begin{figure}[t]
  \centering
  \includegraphics[width=0.95\columnwidth]{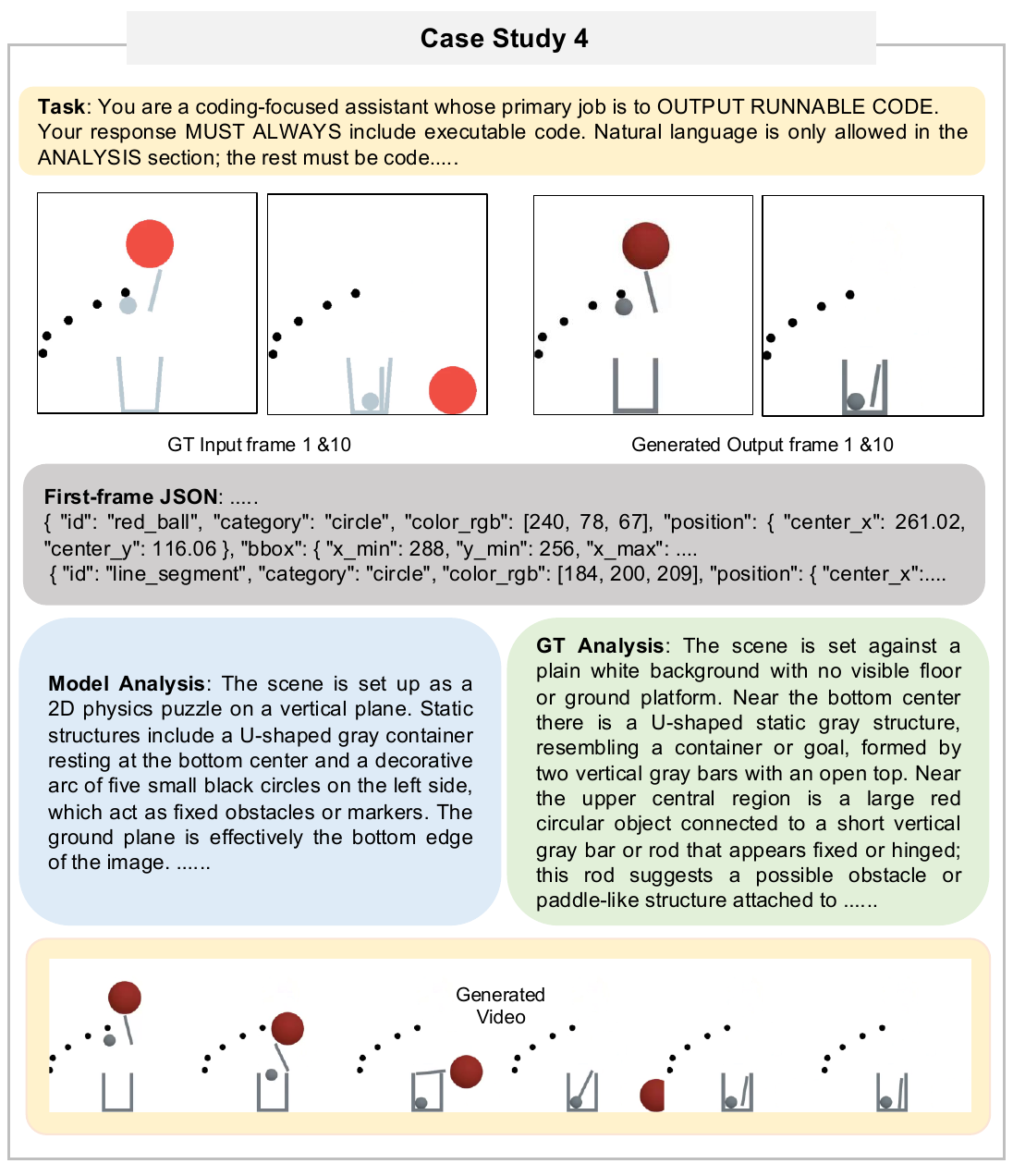}
  \caption{
    A detailed case study (ID 4).
  }
  \label{fig:case_study_4}
\end{figure}
\clearpage

\clearpage
\begin{figure}[t]
  \centering
  \includegraphics[width=0.95\columnwidth]{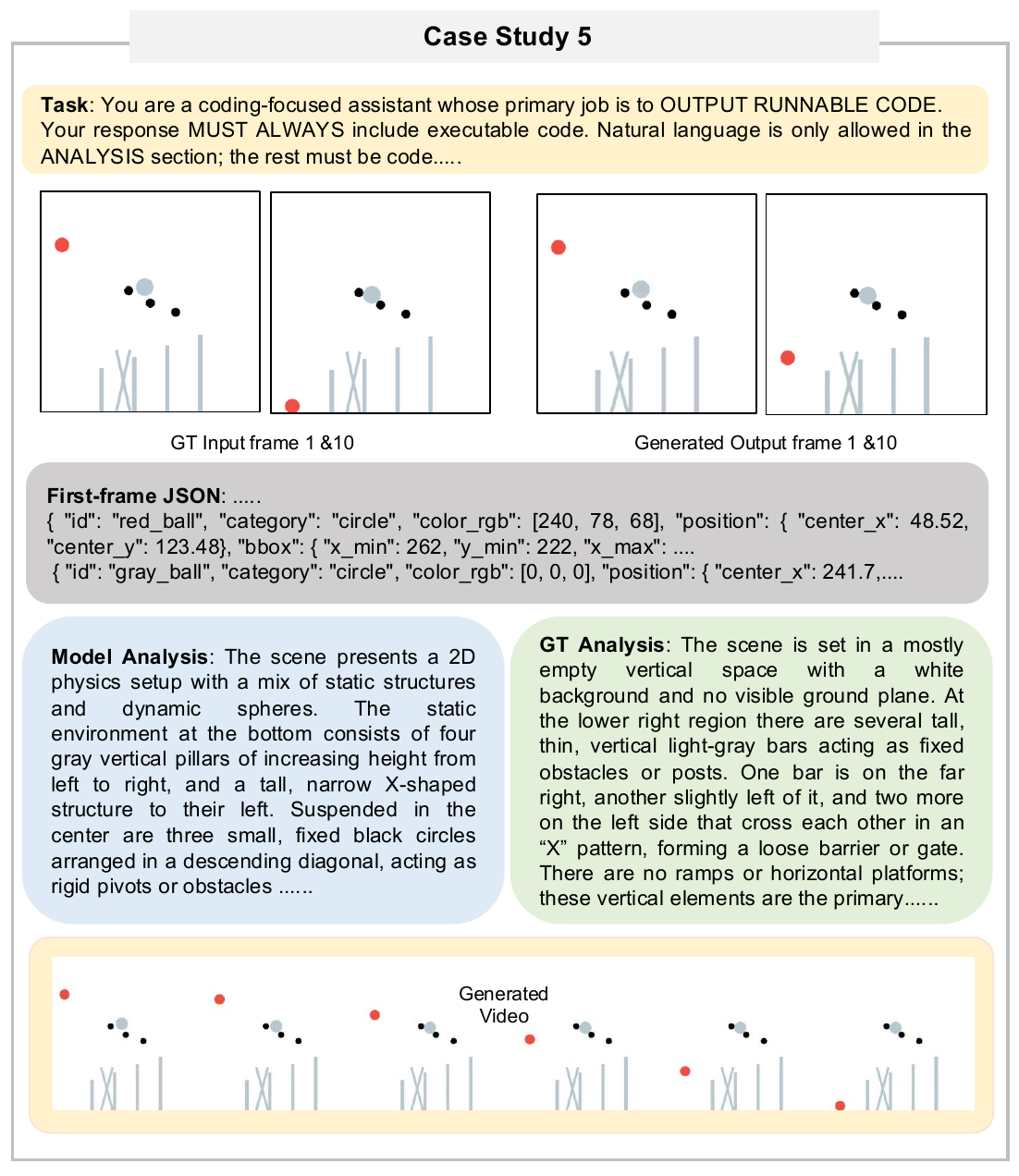}
  \caption{
    A detailed case study (ID 5).
  }
  \label{fig:case_study_5}
\end{figure}
\clearpage

\clearpage
\begin{figure}[t]
  \centering
  \includegraphics[width=0.95\columnwidth]{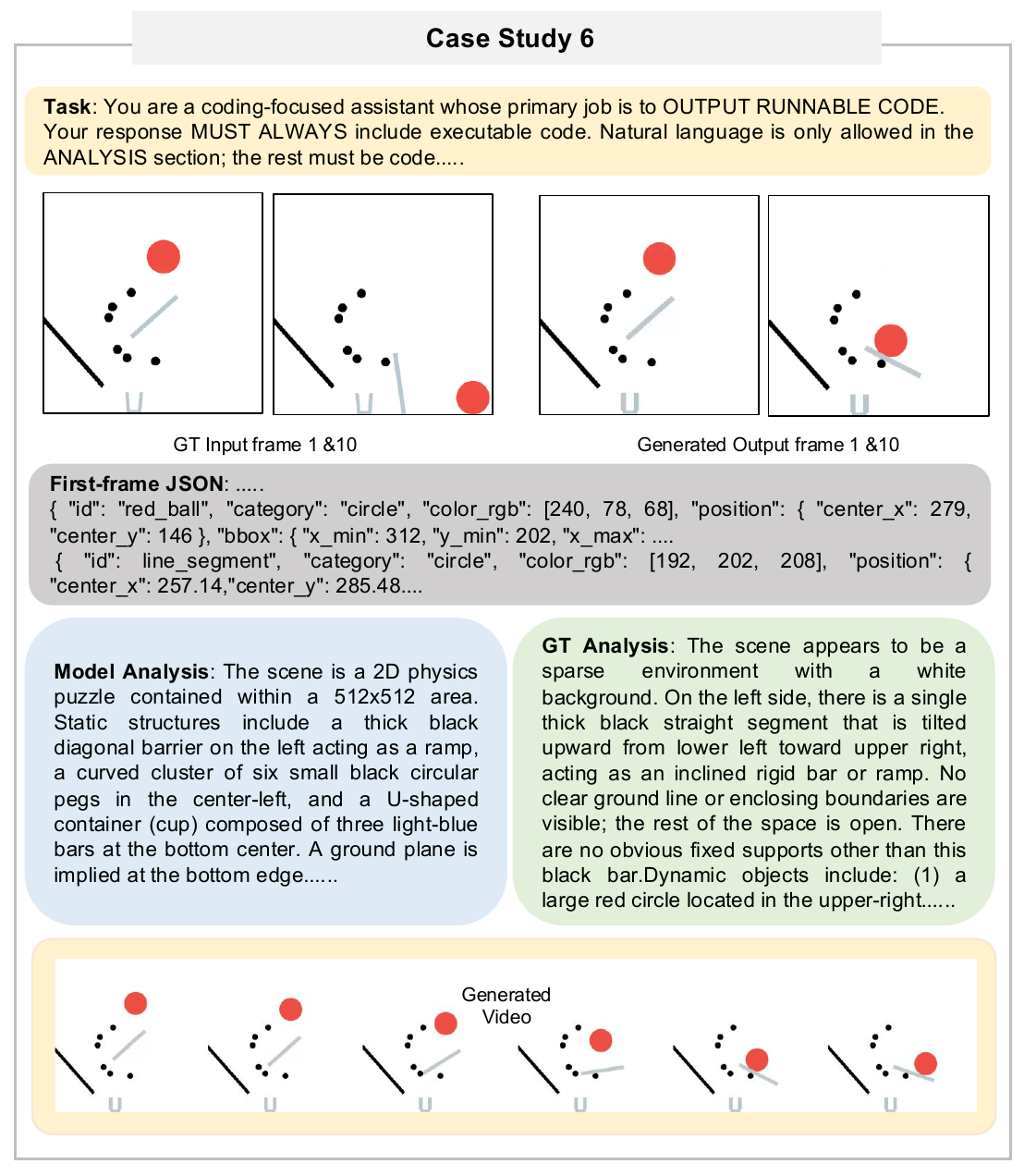}
  \caption{
    A detailed case study (ID 6).
  }
  \label{fig:case_study_6}
\end{figure}
\clearpage

\clearpage
\begin{figure}[t]
  \centering
  \includegraphics[width=0.95\columnwidth]{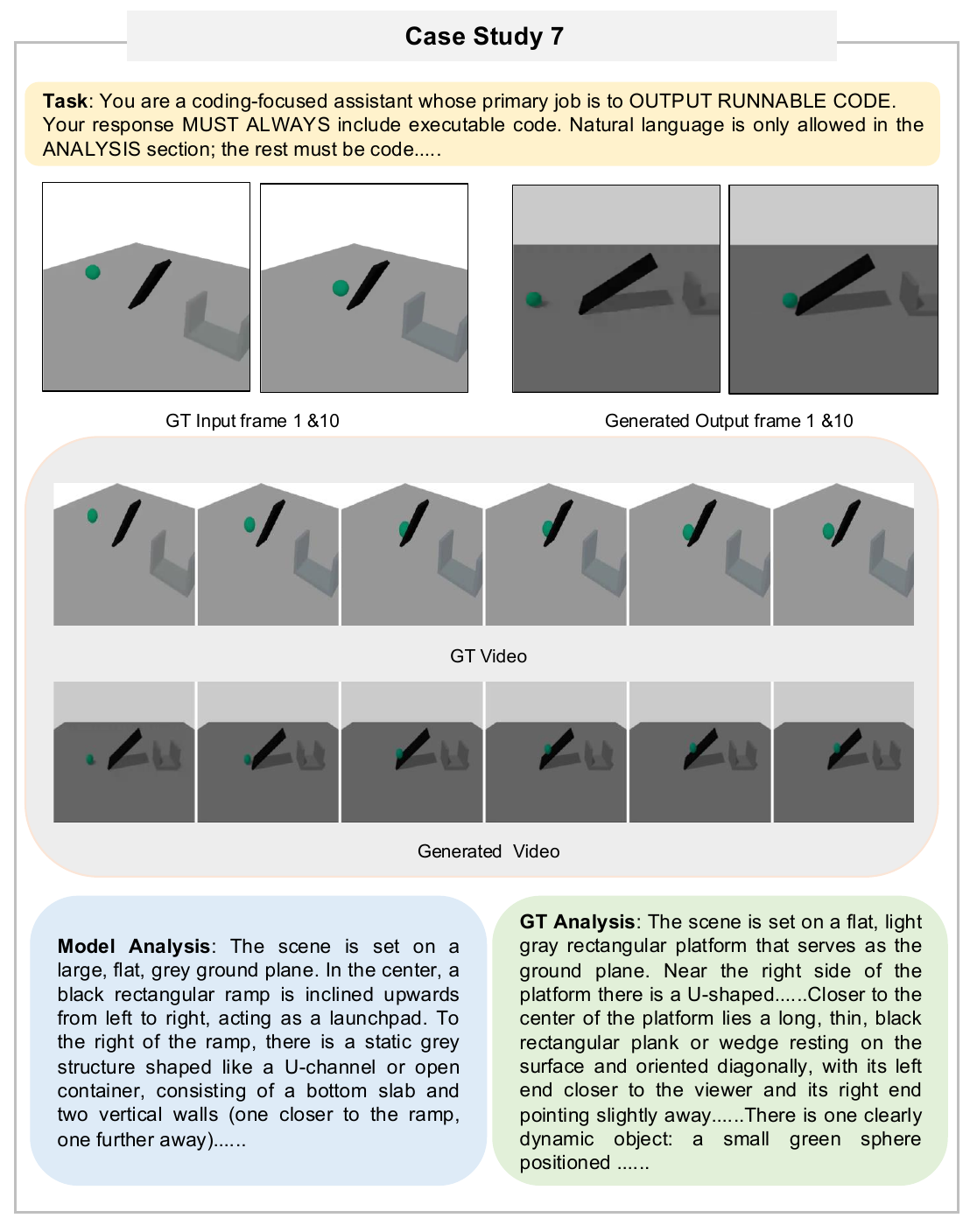}
  \caption{
    A detailed case study (ID 7).
  }
  \label{fig:case_study_7}
\end{figure}
\clearpage

\clearpage
\begin{figure}[t]
  \centering
  \includegraphics[width=0.95\columnwidth]{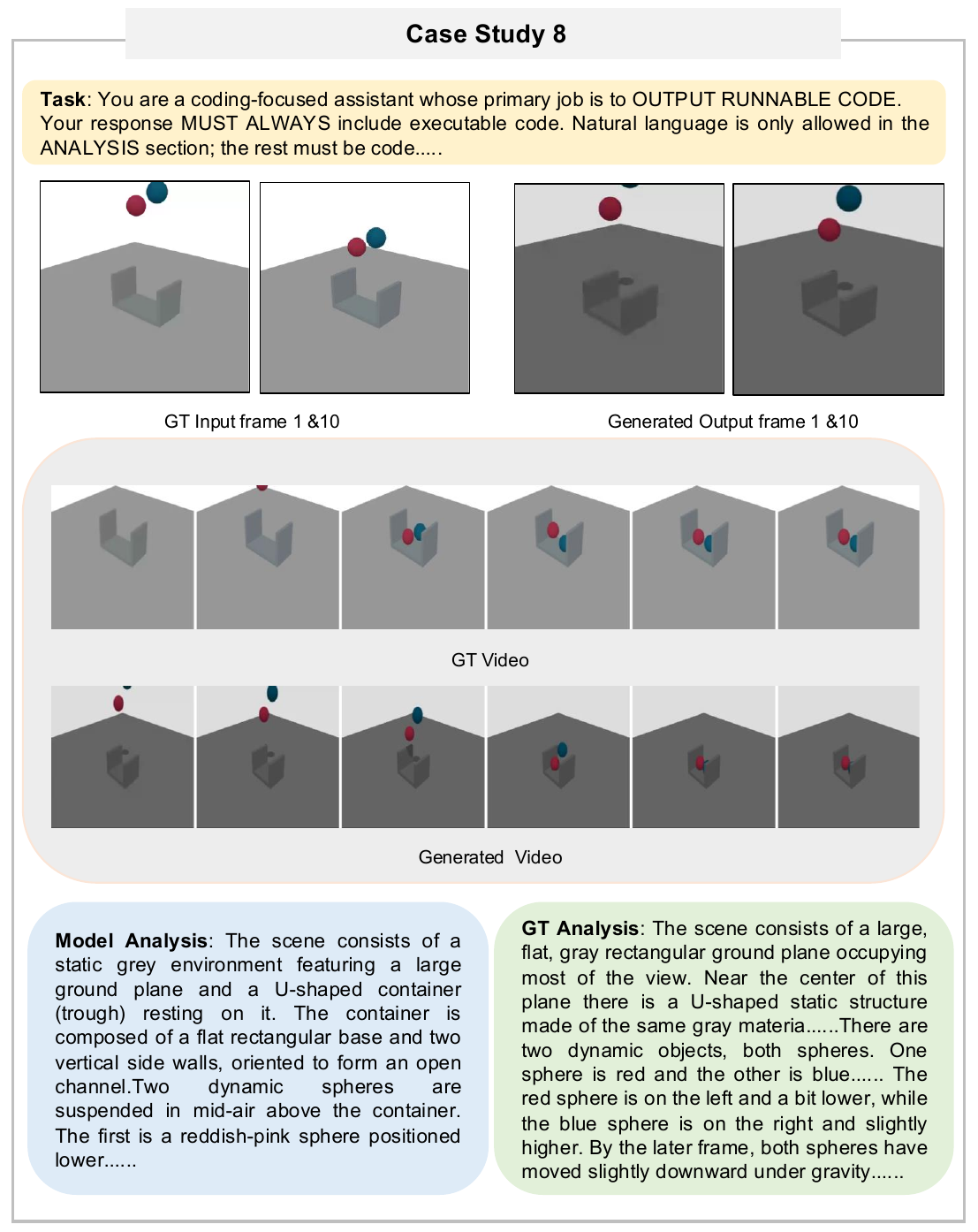}
  \caption{
    A detailed case study (ID 8).
  }
  \label{fig:case_study_8}
\end{figure}
\clearpage

\newpage
\section{Reproducibility Details}
\label{app:implementation}

This appendix documents the reproducibility-critical components of VisPhyWorld:
(i) the prompting protocol used to elicit an executable scene hypothesis,
(ii) the optional detection context format,
(iii) deterministic execution constraints for rendering,
and (iv) robustness protocols that ensure a well-defined evaluation.

\subsection{Prompting Protocol for Scene Hypotheses}
\label{app:prompt-template}

VisPhyWorld uses a single-call prompting protocol that asks the model to
(1) summarize the observed motion between two keyframes and
(2) propose an executable scene hypothesis that reproduces the event.
To ensure comparability across models, we enforce a fixed output format and a
small set of execution constraints (e.g., a single canvas and bounded duration),
which are handled by the renderer (Appendix~\ref{app:renderer-figure}).
The full prompt template is shown in Figure~\ref{fig:visphyworld-llm-prompt}.

\begin{figure*}[!htbp]
\centering
\begin{tcolorbox}[visprompt,title={Scene Analysis \& Code Generation Prompt}]
You are an expert in 2D physics, rigid-body simulation, and JavaScript.

Given two key frames from a short video and an optional list of detected
objects, your goals are:

\textbf{(1) Scene and motion analysis} (3--8 sentences):
\begin{itemize}
  \item Describe the main objects (shapes, colors, approximate sizes) visible in the first frame.
  \item Describe how these objects move between the first and the second frame
        (who moves, who stays still, collisions, stacks that topple, etc.).
  \item Explain the likely physical causes of the motion (gravity, contact forces,
        friction, impulses).
\end{itemize}

\textbf{(2) Simulation code generation:} Produce ONE complete HTML document that:
\begin{itemize}
  \item Imports the required rendering/physics libraries (or uses provided local copies if available).
  \item Creates a 2D-like scene with an orthographic camera.
  \item Adds rigid bodies (balls, boxes, planks) matching the layout of the first frame.
  \item Initializes positions and orientations so that the first rendered frame closely
        matches the first image.
  \item Assigns velocities or impulses consistent with the observed motion between the
        two images.
  \item Runs a physics simulation and renders frames to a single canvas element.
  \item Uses the provided recording helper to export a finite-duration clip.
\end{itemize}

Return your answer in the following format:

\textbf{(A) Analysis section:} plain English paragraphs.

\textbf{(B) Code section:} a single fenced block

\quad\texttt{```html}\\
\quad\texttt{<!DOCTYPE html>} \dots \texttt{</html>}\\
\quad\texttt{```}

Do not include any other Markdown fences or extra HTML documents.
\end{tcolorbox}
\caption{Full multimodal LLM prompt template used by VisPhyWorld for both motion analysis and code generation.}
\label{fig:visphyworld-llm-prompt}
\end{figure*}

\begin{figure*}[!htbp]
\centering
\begin{tcolorbox}[visprompt,title={3D Scene Analysis \& Code Generation Prompt (dataset\_3D)}]
You are an expert in 3D rigid-body physics, Three.js, and JavaScript.

Given two key frames from a short video (rendered from a fixed camera) and an optional list of detected objects, your goals are:

\textbf{(1) Scene and motion analysis} (3--8 sentences):
\begin{itemize}
  \item Describe the major objects and supports visible in the first frame (shapes, colors, approximate sizes, and relative depth if evident).
  \item Describe how these objects move between the first and the second frame, focusing on contacts, impacts, and constraint satisfaction.
  \item Explain the likely physical causes of the motion (gravity, contact forces, friction, impulses).
\end{itemize}

\textbf{(2) Simulation code generation:} Produce ONE complete HTML document that:
\begin{itemize}
  \item Uses Three.js for rendering and Cannon.js for rigid-body simulation.
  \item Creates a \textbf{3D} scene with a fixed \emph{perspective} camera (no camera motion).
  \item Initializes objects and supports so the first rendered frame closely matches the first image.
  \item Assigns velocities or impulses consistent with the observed motion between the two images.
  \item Runs the physics simulation deterministically and renders frames to a single canvas element.
  \item Uses the provided recording helper to export a finite-duration clip.
\end{itemize}

Return your answer in the following format:

\textbf{(A) Analysis section:} plain English paragraphs.

\textbf{(B) Code section:} a single fenced block

\quad\texttt{```html}\\
\quad\texttt{<!DOCTYPE html>} \dots \texttt{</html>}\\
\quad\texttt{```}

Do not include any other Markdown fences or extra HTML documents.
\end{tcolorbox}
\caption{3D prompt variant used for \texttt{dataset\_3D}. It mirrors Figure~\ref{fig:visphyworld-llm-prompt} but switches the execution target from 2D (orthographic) to 3D (fixed perspective) while keeping the output format identical.}
\label{fig:visphyworld-llm-prompt-3d}
\end{figure*}

\subsection{Detection Context $D$}
\label{app:detection-context}

To reduce ambiguity in object discovery and initialization, VisPhyWorld can
optionally provide a structured detection context $D$ for the first keyframe
$I^{\text{start}}$.
$D$ is a per-sample JSON annotation containing a list of objects with coarse
geometry and appearance attributes.
All coordinates are in pixel space with origin at the image top-left
($x$ increases rightward, $y$ increases downward).

\textbf{Schema.}
Each detected object provides:
(i) a unique identifier \texttt{id};
(ii) a coarse \texttt{category} (e.g., \texttt{circle}, \texttt{rectangle},
\texttt{line}, \texttt{u\_shape});
(iii) an RGB color triplet \texttt{color\_rgb};
(iv) a tight bounding box \texttt{bbox} as
\{\texttt{x\_min}, \texttt{y\_min}, \texttt{x\_max}, \texttt{y\_max},
\texttt{width}, \texttt{height}\};
(v) a centroid \texttt{position.center\_x/center\_y};
(vi) a coarse \texttt{size} descriptor (e.g., \texttt{radius\_pixels} for circles,
\texttt{length\_pixels}/\texttt{thickness\_pixels} for bars);
and (vii) an optional \texttt{orientation.angle\_deg} for elongated primitives.

\textbf{Example.}
\begin{verbatim}
{
  "image_size": {"width": 512, "height": 512},
  "coordinate_system": {"origin": "top_left", "x_axis": "to_right"...},
  "objects": [
    {"id":"red_ball","category":"circle","color_rgb":[240,78,70],
     "position":{"center_x":363.6,"center_y":155.2},
     "bbox":{"x_min":348,"y_min":140,"x_max":378,"y_max":172,"width":32...},
     "size":{"radius_pixels":16.5}}
  ]
}
\end{verbatim}

\subsection{VisPhyBench Templates and Stochasticity}
\label{app:visphybench-templates}

VisPhyBench templates are defined as executable PHYRE-style task scripts.
Unlike static assets, each template is instantiated by sampling seeds (e.g., object placements and sizes), so a single rendered snapshot does not capture the full diversity.
We therefore summarize object composition over the full \texttt{sub} split using the detection context $D$ on $I^{\text{start}}$ (see Table~\ref{tab:visphybench-template-categories}).

\begin{table}[!htbp]
  \caption{Object category statistics on VisPhyBench.}
  \centering
  \small
  \setlength{\tabcolsep}{6pt}
  \begin{tabular}{lrrr}
    \toprule
    Category & Scenes (\%) & Scenes (count) & Objects (count) \\
    \midrule
    \texttt{circle} & 100.0 & 191 & 779 \\
    \texttt{line} & 83.2 & 159 & 344 \\
    \texttt{rectangle} & 62.8 & 120 & 321 \\
    \texttt{u\_shape} & 24.6 & 47 & 47 \\
    \texttt{triangle} & 6.3 & 12 & 16 \\
    \texttt{composite\_shape} & 7.3 & 14 & 16 \\
    \bottomrule
  \end{tabular}
  \label{tab:visphybench-template-categories}
\end{table}

\textbf{3D templates.}
In addition to PHYRE-style 2D scripts, we include a set of programmatic 3D templates implemented in Three.js + Cannon.js.
These 3D templates use simple rigid-body primitives (e.g., spheres, boxes, ramps, barriers) under a fixed perspective camera and white background, and are designed to probe depth-aware contacts and occlusions not present in purely 2D scenes.
Because $D$ is defined in 2D pixel space from a first-frame detector, the category statistics above are reported for the 2D portion of the split; for the 3D subset we instead rely on the executable template specification and deterministic rendering protocol (Appendix~\ref{app:renderer-figure}).

\subsection{Deterministic Execution and 2D Constraint}
\label{app:renderer-figure}

VisPhyWorld executes each generated scene hypothesis under a fixed, deterministic
configuration to ensure comparability across models.

\textbf{Canonicalization and validation.}
Raw model outputs may contain extraneous text or malformed markup.
Before execution, we extract the HTML payload (from a fenced \texttt{```html}
block when present, otherwise the outermost \texttt{<html>...</html>} segment),
and canonicalize it into a standard executable template that injects the required
libraries and a trusted recording helper.
We additionally validate basic requirements (e.g., existence of a drawable
canvas and finite numeric states).
Retry and fallback behaviors are described in Appendix~\ref{app:robustness}.

\textbf{Execution contract.}
For each sample, the renderer produces a fixed-length clip $\hat{X}$ at the
reference frame rate and duration associated with that sample.
All runs use a fixed physics time step and a fixed camera configuration; as a
result, variability in $\hat{X}$ is attributable to the generated hypothesis
rather than nondeterministic execution.

\textbf{2D constraint.}
Although the underlying physics engine supports full 3D dynamics, we restrict
motion to a 2D plane by (i) initializing all bodies with $z=0$ and
(ii) projecting the state back to the plane at each simulation step (clamping
out-of-plane position and angular components to zero). This avoids uncontrolled
3D degrees of freedom while preserving rigid-body contact dynamics.

\textbf{3D execution.}
For our 3D subset, we disable the 2D clamping rule and execute full 3D rigid-body dynamics with the same deterministic protocol (fixed physics time step, fixed recording duration, and fixed camera parameters). To preserve comparability across models, we keep the camera static and normalize all rendered videos to match the reference FPS, duration, and resolution of the corresponding ground-truth clip.

\begin{figure*}[!htbp]
\centering
\begin{tcolorbox}[visprompt,title={Deterministic Rendering Protocol (High-Level)}]
\small
Given a model-generated scene hypothesis $C$, the renderer produces the output
clip $\hat{X}$ under a fixed protocol:

\begin{itemize}
  \item \textbf{Parse \& canonicalize:} Extract the HTML payload and wrap it into a
        standard executable template with fixed library versions and a trusted recorder.
  \item \textbf{Validate:} Check minimal execution requirements (e.g., a drawable
        canvas and finite numeric states).
  \item \textbf{Execute deterministically:} Run physics with a fixed time step and
        a fixed orthographic camera, producing frames at the sample's reference FPS.
  \item \textbf{Enforce 2D:} Initialize with $z=0$ and clamp out-of-plane components
        each step.
  \item \textbf{Export:} Record a fixed-duration clip and convert it to a standard
        format for downstream evaluation.
\end{itemize}
\end{tcolorbox}
\caption{High-level deterministic rendering protocol used in VisPhyWorld. Low-level implementation details are included in the released codebase.}
\label{fig:visphyworld-renderer-pipeline}
\end{figure*}

\subsection{Robustness: Automatic Retry and Fallback}
\label{app:robustness}

To handle syntax errors or runtime exceptions in model-generated programs, we
implement a lightweight robustness protocol that ensures evaluation is
well-defined for all samples.

\textbf{Error-conditioned single-step repair.}
If the initial program fails to execute (e.g., syntax error, missing canvas, or
runtime exception), we capture execution diagnostics (e.g., JavaScript console
logs and error traces), summarize them, and provide the summary to the model for
a single repair attempt.

\textbf{Fallback and well-defined evaluation.}
If the repair attempt also fails, we execute a minimal hand-crafted fallback
template (Figure~\ref{fig:fallback-template}) that guarantees a valid canvas and
finite motion. This prevents missing outputs and ensures the evaluation pipeline
does not crash; such samples receive correspondingly poor scores on the metrics.

\textbf{Success criteria.}
We distinguish two notions of success.
\textbf{Model-success} counts a sample as successful only if the model-generated
hypothesis executes and produces a non-empty clip without invoking the fallback.
\textbf{System-success} additionally counts fallback clips as successful, and is
used only to guarantee that the evaluation pipeline is well-defined.
Unless otherwise stated, success rates reported in the main paper use
Model-success.
The 97.7\% success rate in the abstract refers to this model-success criterion
before fallback, aggregated over benchmark runs; fallback clips are used only to
keep metric computation well-defined when generation fails.

\begin{figure*}[htbp]
\centering
\begin{tcolorbox}[visprompt,title={Fallback Template (Simplified Sketch)}]
\small
The fallback template guarantees a valid canvas and finite motion when model
generation fails:

\vspace{0.5em}
\noindent\texttt{<!DOCTYPE html>} \\
\texttt{<html lang="en">} \\
\quad\texttt{<head>} \\
\quad\quad\texttt{<meta charset="UTF-8" />} \\
\quad\quad\texttt{<title>VisPhyWorld Fallback Scene</title>} \\
\quad\quad\texttt{<script src="three.min.js"></script>} \\
\quad\quad\texttt{<script src="cannon.min.js"></script>} \\
\quad\quad\texttt{<script src="recording.js"></script>} \\
\quad\texttt{</head>} \\
\quad\texttt{<body style="margin:0;overflow:hidden;">} \\
\quad\quad\texttt{<canvas id="visphyworld-canvas"></canvas>} \\
\quad\quad\texttt{<script>} \\
\quad\quad\quad\texttt{// Setup renderer, camera, scene, lights} \\
\quad\quad\quad\texttt{// Create a flat ground plane and one spherical body} \\
\quad\quad\quad\texttt{// Run simulation loop and export a finite-duration clip} \\
\quad\quad\texttt{</script>} \\
\quad\texttt{</body>} \\
\texttt{</html>}
\end{tcolorbox}
\caption{High-level structure of the fallback template used when both model attempts fail.}
\label{fig:fallback-template}
\end{figure*}

\section{Evaluation Metrics: Definitions \& Protocols}
\label{app:metrics-details}

This appendix defines the metric families used in the main paper.
All metrics are computed per scene and then averaged over the evaluated split.
Unless otherwise noted, frame-wise metrics are computed after temporal alignment
(Appendix~\ref{app:alignment-defaults}).

\subsection{Default Evaluation Hyperparameters}
\label{app:alignment-defaults}

We report our default evaluation hyperparameters for reproducibility.
Unless otherwise stated, we uniformly sample frames every
\texttt{sample\_every=3} frames for all frame-wise metrics.
For temporal alignment, we use a coarse-to-fine strategy with coarse offset
search up to \texttt{max\_offset=30} (in sampled frames), a stack window
\texttt{window=3}, and offset penalty \texttt{offset\_penalty=0.05}.
The coarse search uses \texttt{downsample=64}, \texttt{top\_k=5}, and
\texttt{max\_samples=16}. When DTW is enabled, we compute frame features using a
$48\times48$ grayscale thumbnail and a step penalty of \texttt{0.005}.

\begin{table*}[htbp]
\caption{Default evaluation hyperparameters used throughout the paper.}
\vspace{3pt}
\centering
\small
\setlength{\tabcolsep}{6pt}
\begin{tabular}{l l}
\toprule
Setting & Value \\
\midrule
Frame sampling & \texttt{sample\_every=3} \\
Coarse offset search & \texttt{max\_offset=30} (sampled frames) \\
Stack refinement window & \texttt{window=3} \\
Offset penalty & \texttt{offset\_penalty=0.05} \\
Coarse downsample & \texttt{downsample=64} \\
Top-$k$ candidates & \texttt{top\_k=5} \\
Max coarse samples & \texttt{max\_samples=16} \\
DTW feature size & $48\times48$ grayscale \\
DTW step penalty & \texttt{0.005} \\
\bottomrule
\end{tabular}

\label{tab:eval-hparams}
\end{table*}

\subsection{Reconstruction \& Perceptual Quality}

\begin{itemize}
    \item \textbf{PSNR and SSIM:} Computed frame-wise between aligned reference and generated videos. SSIM is averaged across the Y channel and RGB channels.
    \item \textbf{LPIPS, FSIM, VSI, DISTS:} Deep and structural perceptual metrics computed on aligned frames. We use the \texttt{piq} library implementation.
\end{itemize}

\subsection{Visual Semantic Consistency}

\begin{itemize}
    \item \textbf{CLIP-Img:} Cosine similarity between CLIP (ViT-B/32) embeddings of reference and generated frames, measuring high-level semantic/layout consistency.
    \item \textbf{DINO Similarity:} Cosine similarity of DINO ViT features, which is more sensitive to object structure and less biased by text supervision than CLIP.
\end{itemize}

\subsection{Text--Video \& Analysis Consistency}

\begin{itemize}
    \item \textbf{CLIP-Cap:} Similarity between the generated motion-analysis text and the generated video frames.
    \item \textbf{Text Metrics (ROUGE, BERTScore):} We compare the generated analysis against an automatically produced reference description of the original video (generated by a strong LLM). This validates whether the model correctly perceives and verbalizes the events in the input video.
\end{itemize}

\subsection{Motion \& Physical Plausibility}

\begin{itemize}
    \item \textbf{RAFT Optical Flow:} We compute End-Point Error (EPE), flow magnitude difference, and angular error between the optical flow fields of the reference and generated videos.
    \item \textbf{Temporal Alignment:} We use a coarse-to-fine alignment strategy (Figure~\ref{fig:temporal-alignment}) combining offset search and Dynamic Time Warping (DTW) to handle temporal shifts before metric computation.
\end{itemize}

\begin{figure}[htbp]
\centering
\begin{tcolorbox}[visprompt,title={Temporal Alignment Procedure}]
\small
\textbf{(1) Coarse Search:} Downsample frames to grayscale vectors. Compute correlation for offsets $\pm 30$ frames. Keep top-$k$ candidates.

\textbf{(2) Stack Refinement:} Build frame stacks ($w=3$). Minimize cost = MSE + 0.5 MAE + 0.1 Angular.

\textbf{(3) DTW:} Run Dynamic Time Warping on low-res features to align variable-speed sequences.
\end{tcolorbox}
\caption{Temporal alignment procedure used before computing frame-wise metrics.}
\label{fig:temporal-alignment}
\end{figure}

\subsection{Subjective Quality (Gemini Judge)}
\label{sec:app-gemini}

We employ \textbf{Gemini-2.5-Pro} as a holistic judge. The prompt (Figure~\ref{fig:gemini-eval-prompt}) asks the model to compare the reference and generated videos and assign a score (1--10) with a justification.
Scores are intentionally conservative: a score near 3--4 indicates that a video
captures some objects or layout but still contains clear motion, contact, or
temporal-consistency errors. We do not rescale these scores.

\begin{figure*}[htbp]
\centering
\begin{tcolorbox}[visprompt,title={Gemini-based Physics \& Video Consistency Prompt}]
You are an expert evaluator of \textbf{physical simulations} and video quality.

Compare the provided \textbf{reference video} (Ground Truth) with the \textbf{generated video}.
Your goal is to determine if the generated video accurately reconstructs the \textbf{physical event} shown in the reference.

Focus on the following dimensions:
\begin{itemize}
  \item \textbf{Physical Plausibility (Crucial):} Do the objects obey rigid-body physics laws (gravity, collisions, friction) in the given setting (2D or 3D)? Are there any ``hallucinations'' such as objects passing through each other (ghosting), floating unnaturally, or failing to move when hit?
  \item \textbf{Motion Consistency:} Does the trajectory, speed, and timing of the movement align with the reference?
  \item \textbf{Scene Semantics:} Are the correct objects (color, shape, count) present in the correct layout?
  \item \textbf{Visual Fidelity:} Overall clarity, ignoring minor rendering style differences if the physics is correct.
\end{itemize}

Return a JSON object with the keys:
\begin{itemize}
  \item \texttt{score}: integer between 1 and 10.
        \begin{itemize}
            \item \textbf{10}: Perfect physical and visual match.
            \item \textbf{1}: Physical laws are violated (e.g., phantom collision, static scene when motion is expected), even if the image looks realistic.
        \end{itemize}
  \item \texttt{justification}: Brief explanation, specifically pointing out any physical violations if present.
\end{itemize}
\end{tcolorbox}
\caption{Prompt template used for Gemini-based evaluation. Note that the prompt is explicitly designed to penalize \textbf{physical violations} (e.g., incorrect collision logic), ensuring the score reflects physical understanding rather than just perceptual similarity.}
\label{fig:gemini-eval-prompt}
\end{figure*}

\paragraph{Human evaluation.}
To calibrate the automatic judge, we additionally conduct a model-blind human
evaluation with six STEM graduate students. Raters compare generated videos
against ground-truth videos in randomized order and assign a 1--5 score. The human ranking shows
the same broad trend as the automatic metrics: strong code-driven models and
Veo-3.1 remain top-tier, while weaker baselines such as SVD and Cosmos are rated
low.

\paragraph{VideoScore2 calibration.}
We also checked VideoScore2~\citep{he2025videoscore2thinkscoregenerative} for the
three code-driven models with available outputs. 

\begin{table*}[htbp]
  \caption{VideoScore2 calibration is reported on a 1--5 scale; higher is better.}
  \vspace{3pt}
  \centering
  \scriptsize
  \setlength{\tabcolsep}{4pt}
  \begin{tabular}{lccccc}
    \toprule
    \textbf{Model} & \textbf{Visual} & \textbf{Temporal} & \textbf{Dynamic} & \textbf{Alignment} & \textbf{Physical} \\
    \midrule
    Claude Sonnet 4.5 & 2.44 & 2.20 & 2.75 & 2.04 & 2.12 \\
    GPT-5 & 2.51 & 2.23 & 2.76 & 2.09 & 2.17 \\
    Gemini-3-Pro & 2.57 & 2.27 & 2.66 & 2.24 & 2.14 \\
    \bottomrule
  \end{tabular}
  \label{tab:app-videoscore2-3d}
\end{table*}

\section{Detailed Experimental Results}
\label{sec:app-metrics}

\subsection{VisPhyBench Difficulty Stratification}
\label{app:VisPhyBench-difficulty}

The main paper reports the difficulty distribution; here we provide additional details on the stratification and split construction. All annotators are graduate students with STEM backgrounds.

\subsection{Per-Scene Distributions and Significance (Sub Split)}
\label{app:distributions-significance}

\begin{figure}[htbp]
  \centering
  \includegraphics[width=\textwidth]{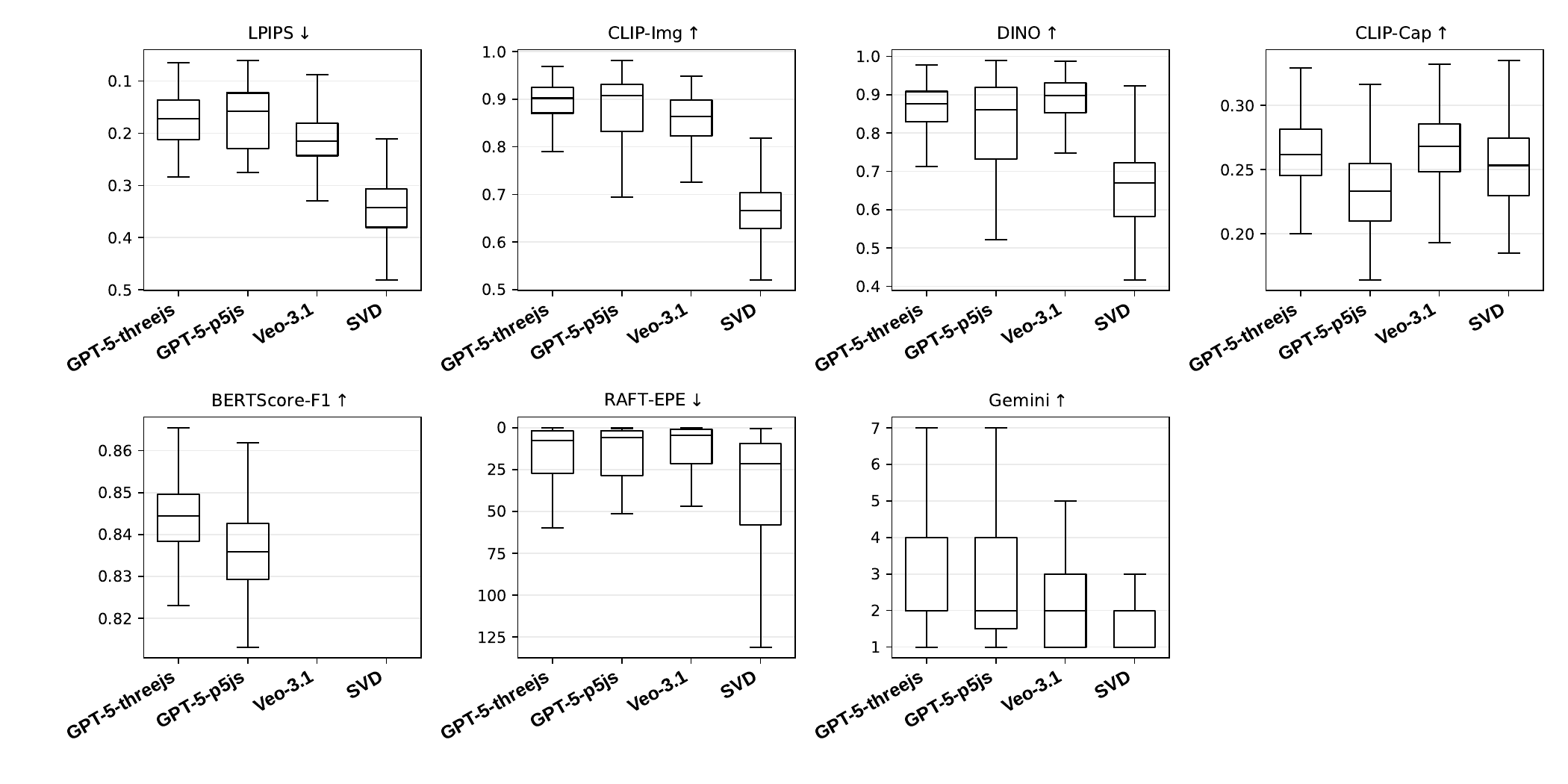}
  \caption{Per-scene boxplot distributions on VisPhyBench for representative metric families (higher is better unless marked $\downarrow$).}
  \label{fig:visphybench-metric-distributions}
\end{figure}

Mean scores can obscure whether improvements are driven by a small subset of scenes.
To address this, we report (i) per-scene metric distributions via boxplots (Figure~\ref{fig:visphybench-metric-distributions}) and (ii) paired bootstrap confidence intervals over per-scene differences (Table~\ref{tab:visphybench-bootstrap-ci}).
We use paired resampling because all methods are evaluated on the same set of scenes ($N=209$), and define ``mean improvement'' so that positive values indicate better performance by VisPhyWorld (GPT-5, threejs), taking metric direction into account ($\uparrow/\downarrow$).

\begin{table*}[htbp]
  \caption{Paired bootstrap confidence intervals (VisPhyBench \texttt{sub}, $N=209$). ``Mean improvement'' is defined so that positive values indicate VisPhyWorld (GPT-5, threejs) performs better (for $\downarrow$ metrics we compute baseline$-$ours; for $\uparrow$ metrics ours$-$baseline).}
  \vspace{3pt}
  \centering
  \small
  \setlength{\tabcolsep}{5pt}
  \begin{tabular}{llcc}
    \toprule
    Metric & Comparison & Mean improvement & 95\% bootstrap CI \\
    \midrule
    LPIPS$\downarrow$ & VisPhyWorld (GPT-5, p5js) & 0.1143 & [0.0743, 0.1567] \\
    LPIPS$\downarrow$ & Veo-3.1 & 0.0365 & [0.0310, 0.0420] \\
    LPIPS$\downarrow$ & SVD (img2vid) & 0.1674 & [0.1581, 0.1764] \\
    CLIP-Img$\uparrow$ & VisPhyWorld (GPT-5, p5js) & 0.0754 & [0.0477, 0.1044] \\
    CLIP-Img$\uparrow$ & Veo-3.1 & 0.0365 & [0.0285, 0.0446] \\
    CLIP-Img$\uparrow$ & SVD (img2vid) & 0.2258 & [0.2159, 0.2355] \\
    DINO$\uparrow$ & VisPhyWorld (GPT-5, p5js) & 0.0957 & [0.0643, 0.1290] \\
    DINO$\uparrow$ & Veo-3.1 & -0.0276 & [-0.0340, -0.0217] \\
    DINO$\uparrow$ & SVD (img2vid) & 0.2036 & [0.1917, 0.2155] \\
    CLIP-Cap$\uparrow$ & VisPhyWorld (GPT-5, p5js) & 0.0299 & [0.0243, 0.0356] \\
    CLIP-Cap$\uparrow$ & Veo-3.1 & -0.0050 & [-0.0098, -0.0002] \\
    CLIP-Cap$\uparrow$ & SVD (img2vid) & 0.0101 & [0.0050, 0.0151] \\
    BERTScore-F1$\uparrow$ & VisPhyWorld (GPT-5, p5js) & 0.0077 & [0.0059, 0.0094] \\
    BERTScore-F1$\uparrow$ & Veo-3.1 & N/A & [N/A, N/A] \\
    BERTScore-F1$\uparrow$ & SVD (img2vid) & N/A & [N/A, N/A] \\
    RAFT-EPE$\downarrow$ & VisPhyWorld (GPT-5, p5js) & 0.6294 & [-0.7743, 2.0695] \\
    RAFT-EPE$\downarrow$ & Veo-3.1 & -0.7078 & [-1.8371, 0.4628] \\
    RAFT-EPE$\downarrow$ & SVD (img2vid) & 11.9706 & [8.8544, 15.1865] \\
    Gemini$\uparrow$ & VisPhyWorld (GPT-5, p5js) & -0.0108 & [-0.5081, 0.5027] \\
    Gemini$\uparrow$ & Veo-3.1 & 0.9153 & [0.4233, 1.3968] \\
    Gemini$\uparrow$ & SVD (img2vid) & 2.0635 & [1.7090, 2.4444] \\
    \bottomrule
  \end{tabular}
  \label{tab:visphybench-bootstrap-ci}
\end{table*}

\subsection{Additional Ablation Studies}
\label{app:additional-ablations}

We report supporting ablation details that isolate whether VisPhyWorld's conclusions depend on renderer self-repair, auxiliary detection context, a single judge model, or access to the later conditioning frame.

\paragraph{Iterative self-repair.}
If the first generation+render attempt fails, VisPhyWorld appends a concise renderer error log tail and the previous attempt to the next LLM call, then retries once.
Table~\ref{tab:ablation-retry-sub} reports success rates with and without this retry mechanism.
This suggests that some failures are correctable surface-level issues such as missing canvas hooks, minor API misuse, or initialization errors rather than irrecoverable physical-scene understanding failures.

\begin{table*}[htbp]
  \caption{Ablation on iterative self-repair (retry) on the VisPhyBench. ``No retry'' counts only samples that succeed on the first generation+render attempt; ``1 retry'' allows one additional attempt with renderer error feedback appended to the prompt.}
  \vspace{3pt}
  \centering
  \scriptsize
  \setlength{\tabcolsep}{4pt}
  \begin{tabular}{lccc}
    \toprule
    Engine & Success (no retry)$\uparrow$ & Success (1 retry)$\uparrow$ & Fixed by retry$\uparrow$ \\
    \midrule
    Three.js & 0.979 & 0.990 & 0.010  \\
    P5.js & 0.853 & 0.979 & 0.126  \\
    \bottomrule
  \end{tabular}
  \label{tab:ablation-retry-sub}
\end{table*}

\paragraph{Detection context.}
Table~\ref{tab:app-ablation-detection} compares matched runs with and without the structured detection context $D$.
Removing $D$ consistently worsens LPIPS and lowers Gemini physical-plausibility scores, while CLIP/DINO/BERTScore move much less.
This indicates that detection context primarily reduces the object discovery and initialization burden; it does not by itself solve the physical dynamics.

\begin{table*}[htbp]
  \caption{Ablation on structured detection context. Lower is better for LPIPS and RAFT-EPE; higher is better for other metrics.}
  \vspace{3pt}
  \centering
  \scriptsize
  \setlength{\tabcolsep}{4pt}
  \begin{tabular}{llccccccc}
    \toprule
    \textbf{Model} & \textbf{Mode} & \textbf{LPIPS$\downarrow$} & \textbf{CLIP-Img$\uparrow$} & \textbf{DINO$\uparrow$} & \textbf{CLIP-Cap$\uparrow$} & \textbf{BERTScore$\uparrow$} & \textbf{RAFT-EPE$\downarrow$} & \textbf{Gemini$\uparrow$} \\
    \midrule
    GPT-5 & with detection & 0.17 & 0.89 & 0.86 & 0.26 & 0.84 & 33.65 & 3.50 \\
    GPT-5 & no detection & 0.29 & 0.88 & 0.81 & 0.28 & 0.85 & 30.55 & 1.98 \\
    Gemini-3-Pro & with detection & 0.14 & 0.90 & 0.84 & 0.26 & 0.85 & 36.20 & 3.80 \\
    Gemini-3-Pro & no detection & 0.21 & 0.90 & 0.84 & 0.25 & 0.85 & 31.77 & 1.84 \\
    Claude Sonnet 4.5 & with detection & 0.16 & 0.90 & 0.83 & 0.26 & 0.85 & 36.20 & 2.39 \\
    Claude Sonnet 4.5 & no detection & 0.27 & 0.89 & 0.82 & 0.25 & 0.86 & 31.80 & 1.65 \\
    \bottomrule
  \end{tabular}
  \label{tab:app-ablation-detection}
\end{table*}

\paragraph{Alternative judge models.}
Tables~\ref{tab:app-ablation-judge-models} and~\ref{tab:app-ablation-judge-correlation} replace the Gemini-2.5-Pro holistic judge with GPT-5.4 and Qwen3-VL-Plus judges.
We selected Qwen3-VL-Plus because it natively supports video input. In contrast, neither the GPT-5 version used in our paper nor the latest GPT-5.4 API currently supports direct video input. Therefore, for GPT-5.4, we uniformly sampled JPEG frames from each video and provided those frames to the model for video-quality evaluation.
Absolute score scales differ across judges, but the weak video baseline SVD-img2vid remains at the bottom, and sample-level correlations are significant across judge pairs.
This reduces the risk that our conclusions are an artifact of a single proprietary evaluator.

\begin{table*}[htbp]
  \caption{Sample-level agreement among judge models. All reported correlations are significant with $p<0.001$.}
  \centering
  \scriptsize
  \setlength{\tabcolsep}{8pt}
  \begin{tabular}{lcc}
    \toprule
    \textbf{Judge pair} & \textbf{Pearson $r$} & \textbf{Spearman $\rho$} \\
    \midrule
    Gemini vs GPT-5.4 & 0.41 & 0.37 \\
    Gemini vs Qwen3-VL & 0.40 & 0.43 \\
    GPT-5.4 vs Qwen3-VL & 0.65 & 0.70 \\
    \bottomrule
  \end{tabular}
  \label{tab:app-ablation-judge-correlation}
\end{table*}

\begin{table*}[htbp]
  \caption{Holistic physical-plausibility scores under different judge models.}
  \vspace{3pt}
  \centering
  \scriptsize
  \setlength{\tabcolsep}{5pt}
  \begin{tabular}{lccc}
    \toprule
    \textbf{Model} & \textbf{Gemini score} & \textbf{GPT-5.4 score} & \textbf{Qwen3-VL score} \\
    \midrule
    Veo-3.1 & 2.56 & 5.09 & 7.82 \\
    Gemini-3-Pro & 3.81 & 3.99 & 6.74 \\
    GPT-5 & 3.53 & 3.33 & 7.12 \\
    GPT-4.1 & 3.04 & 3.33 & 6.84 \\
    Claude-Sonnet-4.5 & 2.50 & 3.60 & 6.52 \\
    Qwen3-VL-Plus & 2.13 & 2.25 & 4.94 \\
    SVD-img2vid & 1.44 & 1.42 & 2.92 \\
    \bottomrule
  \end{tabular}
  \label{tab:app-ablation-judge-models}
\end{table*}

\paragraph{Forecast-only setting.}
Table~\ref{tab:app-ablation-forecast} removes the later frame and conditions only on $I^{\text{start}}$.
Visual metrics and RAFT-EPE change only modestly, but Gemini physical-plausibility scores drop for all tested models.
This suggests that $I^{\text{later}}$ provides useful motion context; without it, models can still produce visually plausible clips but are less likely to recover the correct physical outcome.

\begin{table*}[htbp]
  \caption{Forecast-only ablation. The default setting uses frame$_1$+frame$_{10}$, while forecast-only uses frame$_1$ only.}
  \vspace{3pt}
  \centering
  \scriptsize
  \setlength{\tabcolsep}{4pt}
  \begin{tabular}{llccccccc}
    \toprule
    \textbf{Model} & \textbf{Mode} & \textbf{LPIPS$\downarrow$} & \textbf{CLIP-Img$\uparrow$} & \textbf{DINO$\uparrow$} & \textbf{CLIP-Cap$\uparrow$} & \textbf{BERTScore$\uparrow$} & \textbf{RAFT-EPE$\downarrow$} & \textbf{Gemini$\uparrow$} \\
    \midrule
    GPT-5 & frame$_1$+frame$_{10}$ & 0.30 & 0.87 & 0.81 & 0.27 & 0.85 & 28.23 & 2.70 \\
    GPT-5 & forecast-only & 0.31 & 0.87 & 0.81 & 0.28 & 0.85 & 26.94 & 1.60 \\
    Gemini-3-Pro & frame$_1$+frame$_{10}$ & 0.22 & 0.90 & 0.84 & 0.25 & 0.85 & 30.28 & 2.90 \\
    Gemini-3-Pro & forecast-only & 0.21 & 0.89 & 0.85 & 0.25 & 0.85 & 27.40 & 2.45 \\
    Claude-S4.5 & frame$_1$+frame$_{10}$ & 0.27 & 0.89 & 0.81 & 0.26 & 0.86 & 28.46 & 1.95 \\
    Claude-S4.5 & forecast-only & 0.27 & 0.88 & 0.81 & 0.26 & 0.87 & 28.31 & 1.65 \\
    \bottomrule
  \end{tabular}
  \label{tab:app-ablation-forecast}
\end{table*}

\subsection{Cross-Engine Robustness}
\label{app:cross-engine-robustness}

To separate physical reasoning from backend-specific coding familiarity, we run
GPT-5 using five execution backends:
Three.js, P5.js, SVG, Manim, and Blender. The backends differ in language,
rendering API, and physics support, so large rank changes would indicate a
strong backend confound.
Absolute scores are summarized in Figure~\ref{fig:visphybench-radar}; here we
provide the corresponding rank-correlation matrices. The scene-level ranking for
the primary motion metric is stable: RAFT-EPE has an average pairwise Spearman
correlation of $0.84$ across engines, with all pairwise correlations significant
at $p<0.05$. Appearance and semantic rankings are less stable across engines
(average Spearman correlations $0.47$ for LPIPS and $0.60$ for DINO), which is
expected because rendering style varies more than motion difficulty. Engine choice affects absolute reconstruction quality, but it does not dominate the scene-level physical-difficulty ranking; the main motion-based conclusions remain robust across backends.

\begin{table}[htbp]
  \caption{Cross-engine Spearman correlation matrix on RAFT-EPE. Asterisks indicate $p<0.05$; average pairwise $\rho=0.84$.}
  \centering
  \scriptsize
  \setlength{\tabcolsep}{5pt}
  \begin{tabular}{lccccc}
    \toprule
    \textbf{RAFT-EPE} & \textbf{Three.js} & \textbf{P5.js} & \textbf{SVG} & \textbf{Manim} & \textbf{Blender} \\
    \midrule
    Three.js & 1.00  & --    & --    & --    & --   \\
    P5.js    & 0.70* & 1.00  & --    & --    & --   \\
    SVG      & 0.84* & 0.84* & 1.00  & --    & --   \\
    Manim    & 0.78* & 0.88* & 0.92* & 1.00  & --   \\
    Blender  & 0.75* & 0.89* & 0.88* & 0.88* & 1.00 \\
    \bottomrule
  \end{tabular}
  \label{tab:app-cross-engine-raft}
\end{table}

\begin{table}[htbp]
  \caption{Cross-engine Spearman correlations for visual metrics. Average pairwise correlations are moderate, unlike RAFT-EPE. Asterisks indicate $p<0.05$.}
  \centering
  \scriptsize
  \setlength{\tabcolsep}{5pt}
  \begin{tabular}{llccccc}
    \toprule
    \textbf{Metric} & \textbf{Engine} & \textbf{Three.js} & \textbf{P5.js} & \textbf{SVG} & \textbf{Manim} & \textbf{Blender} \\
    \midrule
    LPIPS & Three.js & 1.00  & --    & --    & --    & --   \\
    LPIPS & P5.js    & 0.43* & 1.00  & --    & --    & --   \\
    LPIPS & SVG      & 0.57* & 0.50* & 1.00  & --    & --   \\
    LPIPS & Manim    & 0.34* & 0.35* & 0.41* & 1.00  & --   \\
    LPIPS & Blender  & 0.52* & 0.57* & 0.55* & 0.46* & 1.00 \\
    \midrule
    DINO & Three.js & 1.00  & --    & --    & --    & --   \\
    DINO & P5.js    & 0.46* & 1.00  & --    & --    & --   \\
    DINO & SVG      & 0.70* & 0.50* & 1.00  & --    & --   \\
    DINO & Manim    & 0.70* & 0.60* & 0.53* & 1.00  & --   \\
    DINO & Blender  & 0.67* & 0.55* & 0.60* & 0.69* & 1.00 \\
    \bottomrule
  \end{tabular}
  \label{tab:app-cross-engine-appearance}
\end{table}

\subsection{PSNR, SSIM, and Execution Success}
Table~\ref{tab:app-psnr-ssim-success} reports frame-wise pixel fidelity and whether each configuration successfully produced valid videos.
This table is separated from the main leaderboard because it focuses on low-level reconstruction and executability rather than the full multi-metric physical-understanding assessment.

\begin{table}[htbp]
\caption{Average PSNR/SSIM and generation success rate on VisPhyBench.}
\centering
\footnotesize
\setlength{\tabcolsep}{4pt}
\begin{tabular}{llccc}
\toprule
Model & Engine & PSNR $\uparrow$ & SSIM $\uparrow$ & Success $\uparrow$ \\
\midrule
GPT-5             & Three.js & 20.54 & 0.9370 & 0.990 \\
GPT-5             & p5.js    & 16.36 & 0.7440 & 0.979 \\
GPT-4.1           & Three.js & 19.74 & 0.9337 & 0.948 \\
GPT-4.1           & p5.js    & 14.83 & 0.6830 & \textbf{1.000} \\
Gemini-3-Pro      & Three.js & \textbf{21.26} & \textbf{0.9445} & 0.957 \\
Gemini-3-Pro      & p5.js    & 15.57 & 0.6943 & 0.963 \\
Claude Sonnet 4.5 & Three.js & 20.75 & 0.9406 & 0.995 \\
Claude Sonnet 4.5 & p5.js    & 15.36 & 0.7160 & \textbf{1.000} \\
Qwen3-VL-Plus     & Three.js & 18.66 & 0.9306 & 0.936 \\
Qwen3-VL-Plus     & p5.js    &  9.14 & 0.4296 & \textbf{1.000} \\
\midrule
SVD (img2vid)     & --       & 14.44 & 0.8802 & \textbf{1.000} \\
Cosmos-Predict2.5-2B & --   & 18.07 & 0.9303 & \textbf{1.000} \\
Veo-3.1           & --       & 20.04 & 0.9354 & \textbf{1.000} \\
\bottomrule
\end{tabular}
\label{tab:app-psnr-ssim-success}
\end{table}

\subsection{Reconstruction \& Perceptual Metrics}
Table~\ref{tab:app-reconst-perceptual} details the pixel-level and perceptual metrics. Gemini-3-Pro consistently achieves the best perceptual scores (LPIPS, FSIM), while Three.js backends generally outperform P5.js.

\begin{table*}[htbp]
    \caption{Detailed breakdown of Reconstruction and Perceptual Metrics.}
    \vspace{3pt}
	    \centering
	    \small
	    \setlength{\tabcolsep}{4pt}
	    \begin{tabular}{lcccccc}
	        \toprule
	        \textbf{Model} & \textbf{PSNR$\uparrow$} & \textbf{SSIM$\uparrow$} & \textbf{LPIPS$\downarrow$} & \textbf{FSIM$\uparrow$} & \textbf{VSI$\uparrow$} & \textbf{DISTS$\downarrow$} \\
	        \midrule
	        VisPhyWorld (GPT-5, threejs)             & 20.54 & 0.9370 & 0.1736 & 0.9014 & 0.8432 & 0.1883 \\
	        VisPhyWorld (GPT-5, p5js)                & 16.36 & 0.7440 & 0.2926 & 0.9105 & 0.8193 & 0.2724 \\
	        VisPhyWorld (GPT-4.1, threejs)           & 19.74 & 0.9337 & 0.1818 & 0.9064 & 0.8309 & 0.2040 \\
	        VisPhyWorld (GPT-4.1, p5js)              & 14.83 & 0.6830 & 0.3520 & 0.8977 & 0.8112 & 0.3348 \\
            VisPhyWorld (Gemini-3-Pro, threejs)      & \textbf{21.26} & \textbf{0.9445} & \textbf{0.1399} & \textbf{0.9225} & \textbf{0.8539} & 0.1859 \\
            VisPhyWorld (Gemini-3-Pro, p5js)         & 15.57 & 0.6943 & 0.3302 & 0.9055 & 0.8220 & 0.3384 \\
            VisPhyWorld (Claude Sonnet 4.5, threejs) & 20.75 & 0.9406 & 0.1602 & 0.9118 & 0.8374 & 0.2001 \\
            VisPhyWorld (Claude Sonnet 4.5, p5js)    & 15.36 & 0.7160 & 0.3250 & 0.9030 & 0.8162 & 0.3109 \\
            VisPhyWorld (Qwen3-VL-Plus, threejs)     & 18.66 & 0.9306 & 0.2207 & 0.8972 & 0.8099 & 0.2373 \\
            VisPhyWorld (Qwen3-VL-Plus, p5js)        & 9.14 & 0.4296 & 0.5478 & 0.8797 & 0.7886 & 0.4396 \\
            \midrule
            SVD (img2vid)                          & 14.44 & 0.8802 & 0.3408 & 0.8239 & 0.7585 & 0.3459 \\
            Cosmos-Predict2.5-2B                   & 18.07 & 0.9303 & 0.2699 & 0.8198 & 0.8194 & 0.3447 \\
            Veo-3.1                                & 20.04 & 0.9354 & 0.2102 & 0.8561 & 0.8586 & \textbf{0.1755} \\
            \bottomrule
        \end{tabular}

    \label{tab:app-reconst-perceptual}
\end{table*}

\subsection{Visual Semantic Consistency}
Table~\ref{tab:app-visual-semantic} compares semantic understanding. GPT-5 and Gemini-3-Pro show strong alignment with the ground truth in terms of CLIP and DINO scores.

\begin{table*}[htbp]
	    \caption{Visual Semantic Consistency Metrics.}
        \vspace{3pt}
	    \centering
	    \small
	    \setlength{\tabcolsep}{4pt}
	    \begin{tabular}{lcc}
	        \toprule
	        \textbf{Model} & \textbf{CLIP-Img$\uparrow$} & \textbf{DINO$\uparrow$} \\
	        \midrule
            VisPhyWorld (GPT-5, threejs)             & 0.8930 & 0.8556 \\
            VisPhyWorld (GPT-5, p5js)                & 0.8134 & 0.7580 \\
            VisPhyWorld (GPT-4.1, threejs)           & 0.8933 & 0.8304 \\
            VisPhyWorld (GPT-4.1, p5js)              & 0.7545 & 0.6786 \\
            VisPhyWorld (Gemini-3-Pro, threejs)      & \textbf{0.8973} & 0.8405 \\
            VisPhyWorld (Gemini-3-Pro, p5js)         & 0.7460 & 0.6721 \\
            VisPhyWorld (Claude Sonnet 4.5, threejs) & 0.8957 & 0.8305 \\
            VisPhyWorld (Claude Sonnet 4.5, p5js)    & 0.7612 & 0.7098 \\
            VisPhyWorld (Qwen3-VL-Plus, threejs)     & 0.8717 & 0.7837 \\
            VisPhyWorld (Qwen3-VL-Plus, p5js)        & 0.6446 & 0.5478 \\
            \midrule
            SVD (img2vid)                          & 0.6677 & 0.6528 \\
            Cosmos-Predict2.5-2B                   & 0.7293 & 0.6516 \\
            Veo-3.1                                & 0.8564 & \textbf{0.8839} \\
	        \bottomrule
	    \end{tabular}

	    \label{tab:app-visual-semantic}
\end{table*}

\subsection{Text \& Physical Consistency}
Table~\ref{tab:text-metrics} and Table~\ref{tab:app-motion-physical} (below) provide the remaining metrics on text analysis quality and physical motion fidelity.

\begin{table*}[htbp]
    \caption{Text--Video and Analysis-Text Consistency Metrics.}
    \vspace{3pt}
    \centering
    \small
    \setlength{\tabcolsep}{4pt}
    \begin{tabular}{lccc}
        \toprule
        \textbf{Model} & \textbf{CLIP-Cap$\uparrow$} & \textbf{ROUGE-L F1$\uparrow$} & \textbf{BERTScore-F1$\uparrow$} \\
        \midrule
        VisPhyWorld (GPT-5, threejs)             & 0.2632 & 0.2186 & 0.8436 \\
        VisPhyWorld (GPT-5, p5js)                & 0.2331 & 0.2057 & 0.8360 \\
        VisPhyWorld (GPT-4.1, threejs)           & 0.2610 & \textbf{0.2383} & \textbf{0.8522} \\
        VisPhyWorld (GPT-4.1, p5js)              & 0.2192 & 0.1689 & 0.8253 \\
        VisPhyWorld (Gemini-3-Pro, threejs)      & 0.2567 & 0.2141 & 0.8460 \\
        VisPhyWorld (Gemini-3-Pro, p5js)         & 0.2184 & 0.1886 & 0.8396 \\
        VisPhyWorld (Claude Sonnet 4.5, threejs) & 0.2588 & 0.2168 & 0.8468 \\
        VisPhyWorld (Claude Sonnet 4.5, p5js)    & 0.2177 & 0.1599 & 0.8224 \\
        VisPhyWorld (Qwen3-VL-Plus, threejs)     & \textbf{0.2650} & 0.2022 & 0.8466 \\
        VisPhyWorld (Qwen3-VL-Plus, p5js)        & 0.2032 & 0.1733 & 0.8358 \\
        \midrule
        SVD (img2vid)                          & 0.2533 & -- & -- \\
        Cosmos-Predict2.5-2B                   & --     & -- & -- \\
        Veo-3.1                                & \textbf{0.2681} & -- & -- \\
        \bottomrule
    \end{tabular}

\label{tab:text-metrics}
\end{table*}

\begin{table*}[htbp]
    \caption{Motion and Physical Plausibility Metrics (Selected columns).}
    \vspace{3pt}
    \centering
    \small
    \setlength{\tabcolsep}{4pt}
    \begin{tabular}{lccc}
        \toprule
        \textbf{Model} & \textbf{RAFT-EPE$\downarrow$} & \textbf{RAFT-Angle$\downarrow$} & \textbf{Align-Err$\downarrow$} \\
        \midrule
        VisPhyWorld (GPT-5, threejs)             & 33.6473 & 68.5500 & 0.0210 \\
        VisPhyWorld (GPT-5, p5js)                & 34.3433 & 75.8555 & 0.0279 \\
        VisPhyWorld (GPT-4.1, threejs)           & 33.7110 & 67.7974 & 0.0249 \\
        VisPhyWorld (GPT-4.1, p5js)              & 37.6993 & 82.9492 & 0.0397 \\
        VisPhyWorld (Gemini-3-Pro, threejs)      & 36.2030 & \textbf{62.4494} & 0.0192 \\
        VisPhyWorld (Gemini-3-Pro, p5js)         & 33.1013 & 81.5723 & \textbf{0.0184} \\
        VisPhyWorld (Claude Sonnet 4.5, threejs) & 36.1985 & 71.7979 & 0.0210 \\
        VisPhyWorld (Claude Sonnet 4.5, p5js)    & 34.1425 & 78.2841 & 0.0277 \\
        VisPhyWorld (Qwen3-VL-Plus, threejs)     & 35.0493 & 75.6650 & 0.0350 \\
        VisPhyWorld (Qwen3-VL-Plus, p5js)        & \textbf{20.8187} & 80.7413 & 0.8567 \\
        \midrule
        SVD (img2vid)                          & 45.4606 & 84.7314 & 0.0746 \\
        Cosmos-Predict2.5-2B                   & 33.1221 & 80.1338 & 0.0455 \\
        Veo-3.1                                & 32.7145 & 77.0550 & 0.0193 \\
        \bottomrule
    \end{tabular}

    \label{tab:app-motion-physical}
\end{table*}


\end{document}